\documentclass{article}

\usepackage[accepted]{icml2024}
\usepackage[colorlinks=true,allcolors=perfblue]{hyperref} 
\usepackage{microtype}
\usepackage{graphicx}
\usepackage{booktabs} %
\usepackage{url}
\usepackage{ dsfont }
\usepackage{graphicx}
\usepackage{amsmath}
\usepackage{amsfonts}
\usepackage{caption}
\usepackage{subcaption}

\usepackage[customcolors]{hf-tikz}
\definecolor{expert}{HTML}{008000}
\definecolor{error}{HTML}{f96565}
\definecolor{learner}{HTML}{F79646}
\definecolor{perfblue}{RGB}{64, 114, 175}

\usepackage{transparent}
\newcommand{\algcommentlight}[1]{\textcolor{perfblue}{\transparent{0.8}\small{\texttt{\textbf{//\hspace{2pt}#1}}}}}

\usepackage{amsmath}
\usepackage{amssymb}
\usepackage{mathtools}
\usepackage{amsthm}
 
\usepackage[capitalize,noabbrev]{cleveref}
\newcommand{\acro}{\texttt{EvIL}}
\theoremstyle{plain}

\theoremstyle{definition}

\theoremstyle{remark}

\usepackage[textsize=tiny]{todonotes}

\icmltitlerunning{EvIL: Evolution Strategies for Generalisable Imitation Learning
}

\begin{document}

\twocolumn[
\icmltitle{EvIL: Evolution Strategies for Generalisable Imitation Learning}

\icmlsetsymbol{equal}{*}

\begin{icmlauthorlist}
\icmlauthor{Silvia Sapora}{ox}
\icmlauthor{Gokul Swamy}{cm}
\icmlauthor{Chris Lu}{ox}
\icmlauthor{Yee Whye Teh}{ox}
\icmlauthor{Jakob Nicolaus Foerster}{ox}
\end{icmlauthorlist}

\icmlaffiliation{ox}{University of Oxford, UK}
\icmlaffiliation{cm}{Carnegie Mellon University, USA}

\icmlcorrespondingauthor{Silvia Sapora}{silvia.sapora@gmail.com}

\icmlkeywords{Machine Learning, ICML, IRL, Inverse Reinforcement Learning, RL, Reinforcement Learning, Shaping, Potential-Based Shaping, Evolution Strategies}

\vskip 0.3in
]

\printAffiliationsAndNotice{} %

\begin{abstract}
Often times in imitation learning (IL), the environment we collect expert demonstrations in and the environment we want to deploy our learned policy in aren't exactly the same (e.g. demonstrations collected in simulation but deployment in the real world). Compared to policy-centric approaches to IL like behavioural cloning, reward-centric approaches like \textit{inverse reinforcement learning} (IRL) often better replicate expert behaviour in new environments. This transfer is usually performed by optimising the recovered reward under the dynamics of the target environment. However, \textit{(a)} we find that modern deep IL algorithms frequently recover rewards which induce policies far weaker than the expert, \textit{even in the same environment the demonstrations were collected in}. Furthermore, \textit{(b)} these rewards are often quite poorly shaped, necessitating extensive environment interaction to optimise effectively. We provide simple and scalable fixes to both of these concerns. For \textit{(a)}, we find that \textit{reward model ensembles} combined with a slightly different training objective significantly improves re-training and transfer performance. For \textit{(b)}, we propose a novel \textit{evolution-strategies} based method (\acro{}) to optimise for a reward-shaping term that speeds up re-training in the target environment, closing a gap left open by the classical theory of IRL. On a suite of continuous control tasks, we are able to re-train policies in target (and source) environments more interaction-efficiently than prior work.
Our code is open-sourced at \textbf{\texttt{\url{https://github.com/SilviaSapora/evil}}}.

\end{abstract}

\begin{figure}[t]
    \centering
    \includegraphics[width=0.4\textwidth]{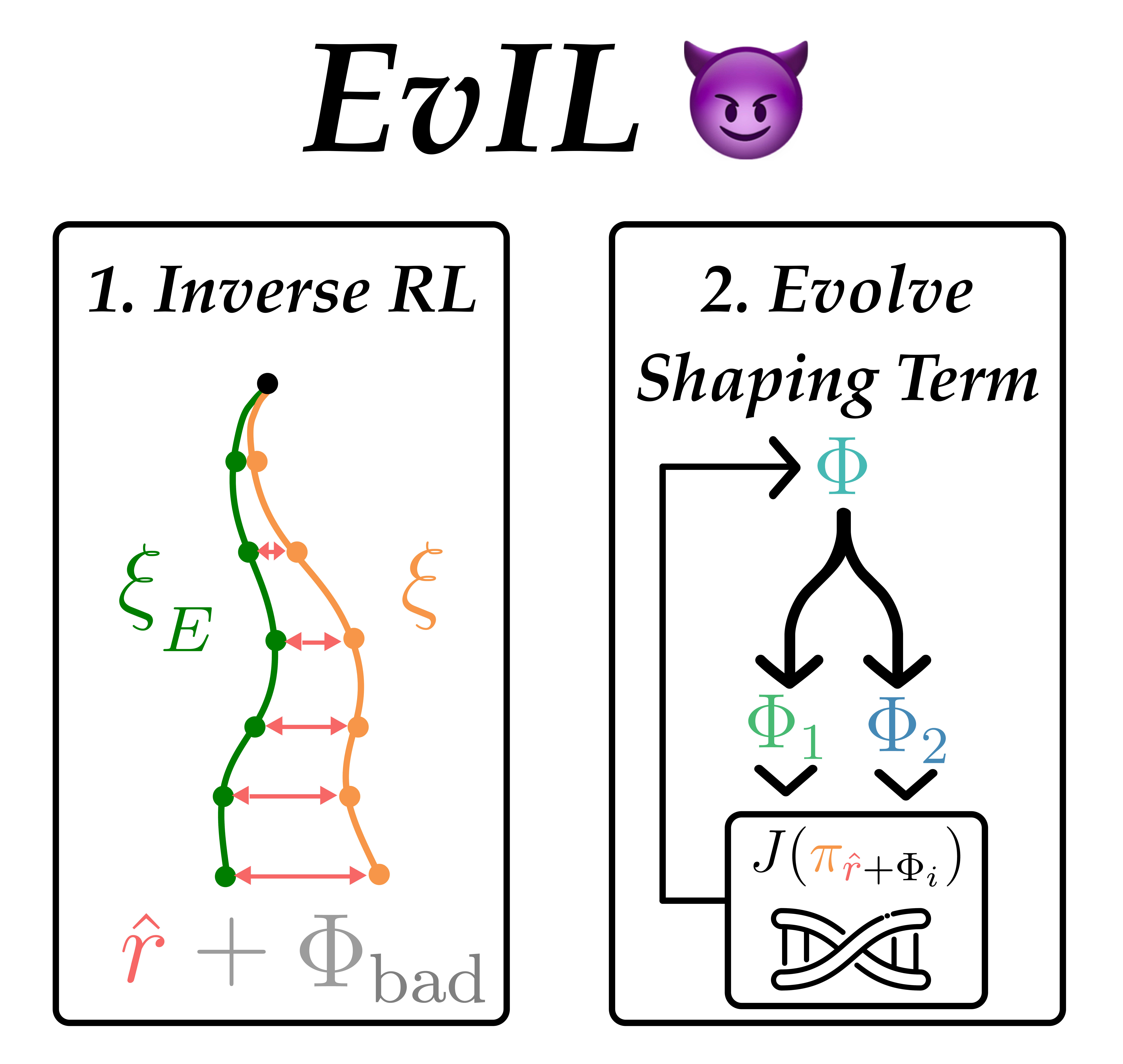}
    \caption{A long-standing concern with standard inverse reinforcement (IRL) algorithms is that they can recover poorly shaped reward functions as the moment-matching loss is \textit{invariant} to potential-based shaping terms \citep{ng2000algorithms}. We propose using \textit{evolution strategies} to learn a shaping term on top of the reward recovered by inverse RL via \textit{directly} optimising for efficient re-training. We are able to re-train policies in both source and unseen target environments more sample-efficiently than the prior work.}
\end{figure}
\section{Introduction}
Consider the problem of predicting driver behaviour across a network of roads. Let's say there are $|\mathcal{S}|$ nodes in our routing graph and $|\mathcal{A}|$ roads going out of each node. If we simply wanted to predict the path usually taken between some start node $s_1$ and some goal node $s_2$, we could collect data of drivers navigating between these two points and learn a policy by regressing from their states to their actions, an approach known as behavioural cloning (BC, \citet{pomerleau1988alvinn}). As we'd need to learn an action at each state, this would require learning $|\mathcal{S}||\mathcal{A}|$ parameters. Now, let's say we wanted to learn the path taken between the same start node $s_1$ and some new goal node $s_3$. We'd need to repeat this entire procedure once again. Thus, to learn how to navigate from one goal to all destinations, a policy-centric approach like BC would need $|\mathcal{S}|^2|\mathcal{A}|$ parameters. In contrast, a reward-centric approach to imitation like \textit{inverse reinforcement learning} (IRL, \citet{ziebart2008maximum}) would simply require learning the $|\mathcal{S}||\mathcal{A}|$ parameters of the underlying reward function from expert data and then could be optimised to find the shortest path between any two nodes in the graph. This ability to learn a \textit{compact} representation that generalises well across multiple tasks / environments was one of the key reasons IRL was developed in the first place. Indeed, as \citet{ng2000algorithms} put it, \textit{``the entire field of reinforcement learning is founded on the presupposition that the reward function, rather than the policy is the most succinct, robust, and transferable definition of the task''}. This intuition remains true in the modern day, perhaps explaining why robust real-world mapping services like Google Maps are build on top of the bedrock of IRL \citep{barnes2023massively}.

If we look at classical IRL algorithms like Maximum Entropy IRL \citep{ziebart2008maximum} or LEARCH \citep{ratliff2009learning}, they usually have a double loop structure. In the outer loop, one trains a discriminator to solve a classification problem between learner and expert trajectories. The discriminator is then used as a reward function in the inner loop, where the learner optimises this adversarially chosen reward function via reinforcement learning.
Observe that because we are actually optimising the reward function to completion in each inner loop iteration, we have reason to believe we will be able to retrain effectively from scratch, a point borne out in practice across a wide variety of disciplines (e.g.\ in robotics \citep{silver2010learning, ratliff2009learning, kolter2008control, ng2006autonomous, zucker2011optimization}, computer vision \citep{kitani2012activity}, and human-computer interaction \citep{ziebart2008navigate, ziebart2012probabilistic}).

More modern approaches to IRL like GAIL \citep{ho2016generative} keep the double loop structure but perform a small amount of policy optimisation in each inner loop iteration, rather than completely resolving the RL problem. 
While this means we expend less environment interactions in each inner loop iteration, it also means that we no longer can ensure that retraining from scratch on the recovered reward function will guarantee performance similar to that of the expert. Beyond being merely a theoretical concern, we find that the rewards recovered by modern deep IL algorithms preclude such \textbf{\textit{effective}} retraining, \textit{even in the environment the expert demonstrations were collected in.} This raises one our work's key questions: \textbf{\textit{how can we preserve the computational benefits of modern IRL while matching the re-training ability of classical IRL?}}

However, even the rewards recovered by classical IRL methods suffer from a weakness: there is no need for them to be nicely shaped, which can preclude \textbf{\textit{efficient}} retraining. In fact, later in the same paper, \citet{ng2000algorithms} raise the following question: \textit{``shaping rewards can produce reward functions that make it dramatically easier to learn a solution to an MDP, without affecting optimality. Can we design IRL algorithms that recover `easy' rewards?''} As we discuss in greater detail in the following sections, classical IRL algorithms are entirely agnostic to reward shaping, as they assume we have the ability to perfectly compute the optimal policy at each iteration and use a loss function for updating the reward estimate that is invariant to shaping terms. Thus, the classical theory of IRL doesn't provide any clear answers for how to make re-training more interaction-efficient.

We provide a method for efficient and effective retraining in IRL that scales to modern deep learning architectures and algorithms. First, \textbf{we find that a combination of \textit{reward model ensembles}, random policy resets, and a more stable loss function allows for more \textit{effective} retraining} that enables the agent to more closely match expert performance. For the second, \textbf{we introduce a novel \textit{evolution-strategies} based method that \textit{directly} optimises for \textit{efficiency} in re-training}, beating out classical value function-based approaches to reward shaping. We refer to the combination of these two techniques as \acro{}: \textit{\textbf{an algorithm that allows effective and efficient re-training in novel environments.}}

More explicitly, our contributions are four-fold:

\noindent \textbf{1. We empirically demonstrate that rewards recovered by conventional IRL algorithms consistently fail in producing optimal agents when employed to retrain an agent from scratch}. This holds true even when the retraining occurs within the same environment where the data was initially collected, differing from the picture in theory.

\noindent \textbf{2. We provide a suite of adjustments that make retraining based on the reward recovered by IRL more effective}. Our adjustments are simple to implemented require minimal additional environment interactions, preserving the efficiency of more modern approaches to IRL.

\noindent \textbf{3. We introduce a novel evolution-strategies based approach to potential-based reward shaping that allows for efficient retraining}. Rather than classical heuristic approaches that attempt to uniformly approximate the value function of the optimal policy, we instead use zeroth-order optimisation to \textit{directly} optimise for interaction-efficient retraining. We also show a speedup in vanilla RL.

\noindent \textbf{4. We perform extensive experimental evaluation of our proposed method across a suite of continuous control tasks and find that it leads to significantly more efficient and effective retraining in source and target environments than prior work}. 
We show that we can combine both of these techniques to reap the benefits of efficient and effective retraining even in environments that are markedly different than what the demonstrations were collected in.

We begin with a discussion of related work.

\section{Related Work} 

\noindent \textbf{Inverse Reinforcement Learning}. IRL is commonly framed as a two-player zero-sum game between a policy player and a reward function player \citep{syed2007game, ho2016generative, swamy2021moments}. Intuitively, the reward function player tries to pick out differences between the current learner policy and the expert demonstration, while the policy player attempts to maximise this reward function to move closer to expert behaviour. As pointed out by \citet{finn2016connection}, this setup is effectively a GAN \citep{goodfellow2014generative} in the space of trajectories. %

IRL algorithms can be categorised into two flavours: \textit{primal} and \textit{dual} \citep{swamy2021moments}. In a primal algorithm, one follows a \textit{no-regret} strategy for both the policy and reward player. Practically, this corresponds to taking a small step on the reward function before performing a small amount of policy optimisation. Most modern deep IL algorithms like GAIL \citep{ho2016generative} and AIRL \cite{fu2018learning} follow this approach. In contrast, the classical approaches to IRL (e.g. MaxEnt IRL \citep{ziebart2008maximum}, LEARCH \cite{ziebart2008maximum}, \citet{abbeel2004apprenticeship}) are of the \textit{dual} flavour: one follows a no-regret strategy for reward selection against a \textit{best response} via RL for policy selection. Practically, this corresponds to optimising the reward to completion at each iteration. For example, in MaxEnt IRL, the best response is performed via soft value iteration. From the perspective of interaction efficiency, primal methods are preferable to dual methods due to lesser interaction in their inner loop. However, from the perspective of ensuring effective retraining (more formally, from the perspective of \textit{best-iterate} convergence), dual methods are preferable due to their explicit retraining at each iteration. Our goal is to achieve the best of both worlds: we seek to preserve the retraining ability of classical dual methods while leveraging the interaction efficiency of modern primal methods.

\noindent \textbf{Efficient Inverse RL.} Various authors have attempted to improve IRL, both in terms of \textit{sample efficiency} (number of expert demonstrations required) and \textit{interaction efficiency}. \citet{swamy2022minimax} provide the minimax-optimal algorithm for IRL in terms of sample efficiency -- this is complimentary to the interaction efficiency we focus on in our work. \citet{swamy2023inverse} provide the first general $\mathsf{poly}(H)$ interaction complexity algorithm for IRL. In contrast, we focus on efficient \textit{retraining}. Furthermore, we can easily combine our approach with theirs as they are orthogonal -- they focus on reducing exploration during policy search under an arbitrary reward function while we provide a method to learn well-shaped reward functions that can be used by an arbitrary downstream policy optimisation procedure. The same can be said for the \textit{hybrid RL} based approach of \citet{ren2023hyrbid} and the BC regularisation suggested by \citet{tiapkin2023regularized}.

\noindent \textbf{Reward Shaping.} Starting with the seminal work of \citet{ng1999policy}, reward shaping has emerged as a critical component of reducing the interaction complexity of reinforcement learning methods \citep{Jaderberg_2019, Wu2017Doom}. As we discuss further below, reward shaping is often thought of as reducing the \textit{effective horizon} of the planning problem. In the search-based planning literature, this is often referred to as \textit{$A^{\star}$ search} \citep{likhachev2003ara, likhachev2005anytime}. Even in the era of deep RL, the effective horizon of a problem has remained as an accurate predictor of the performance of policy optimisation algorithms \citep{laidlaw2023bridging}. From this perspective, the optimal shaping term would be the value function of the optimal policy -- the greedy policy would then be the optimal policy. However, ensuring that we've learned a value function that closely approximates the true optimal value function everywhere in the state space (more formally, an \textit{admissible} heuristic \citep{russell2010artificial}), is a rather tall order, outside of small tabular problems like those considered in \citet{cooke2023toward}. We find that in practice, using the critic of a strong policy as a learned shaping term improves interaction efficiency, but that our approach is able to significantly out-perform this baseline. We hypothesise this is because a learned critic is likely only accurate on states the optimal policy visits, while our method directly optimises for reducing training time by looking for a shaping term that provides signal over the entire course of training, including on the state distribution of the initial weak policy. 

\noindent \textbf{Evolution Strategies for RL.} Recent work leverages JAX-based \citep{jax2018github} hardware acceleration for RL to massively parallelise  training \citep{lu2022discovered}. This has been used for large-scale multi-agent learning \citep{rutherford2023jaxmarl} and rapid population-based training \citep{flajolet2022fast}. Another, more related, line of work leverages these techniques to perform evolution-based \citep{salimans2017evolution} bi-level optimisation. For example, \citet{lu2022discovered, jackson2023discovering} \textit{evolve} surrogate objective functions to effectively discover new, performant reinforcement learning algorithms. \citet{lu2023adversarial, lupu2024behaviour} instead evolve and analyse adversarial environment features and data that influence long-term RL learning behaviour whilst \citet{khan2023scaling, lu2022model} do the same for adversarial policies. Thus, Evolution Strategies (ES) are well-suited for long-horizon and noisy bi-level optimisation tasks, such as those involving an RL inner loop. We apply a similar evolution-based approach for reward shaping.

\section{Background} 
We provide a quick overview of relevant background.

\subsection{Inverse Reinforcement Learning as Game Solving}
\label{subsec:irlgame}
We consider a finite-horizon Markov Decision Process (MDP) \citep{puterman2014markov} parameterized by $\langle \mathcal{S}, \mathcal{A}, \mathcal{T}, H \rangle$ where $\mathcal{S}$, $\mathcal{A}$ are the state and action spaces, $\mathcal{T}: \mathcal{S} \times \mathcal{A}\rightarrow \Delta(\mathcal{S})$ is the transition operator, and $H$ is the horizon. In imitation learning, we see trajectories generated by an expert policy $\pi_E: \mathcal{S} \rightarrow \Delta(\mathcal{A})$, but do not know the reward function. Our goal is to find a reward function that, when optimised, recovers a policy with similar behaviour to that of the expert. We cast this problem as a zero-sum game between a policy player and an adversary that tries to penalise any difference between the expert and learner policies \cite{syed2007game, ho2016generative, swamy2021moments}. More formally, we optimise over policies $\pi: \mathcal{S} \rightarrow \Delta(\mathcal{A}) \in \Pi$ and reward functions $f: \mathcal{S} \times \mathcal{A} \rightarrow [-1, 1] \in \mathcal{F}_r$, where we assume that both $\Pi$ and $\mathcal{F}_r$ are convex and compact so Sion's minimax theorem holds. We use $\xi = (s_1,a_1,r_1,\ldots, s_H,a_H,r_H)$ to denote the trajectory generated by some policy. 
Using $J(\pi, \hat{r}) = \mathbb{E}_{\xi \sim \pi}[\sum_{h=1}^H \hat{r}(s_h, a_h)]$  to denote the value of policy $\pi$ under reward function $\hat{r}$, we have the following:
\begin{equation}
\min_{\pi \in \Pi} \max_{f \in \mathcal{F}_r} J(\pi_E, f) - J(\pi, f). \label{eq:irl-game}
\end{equation}

\subsection{Potential-Based Reward Shaping}
Potential-based reward shaping \citep{ng1999policy} is a technique for speeding up policy optimisation by reducing the effective planning horizon of the problem which guarantees preserving policy optimality. More formally, we define a potential function $\Phi: \mathcal{S} \rightarrow \mathds{R}$ that assigns real values to states $s \in \mathcal{S}$ in the environment. The potential-based shaping term $F: \mathcal{S} \times \mathcal{S} \rightarrow \mathds{R}$ is then defined as the \textit{change} in the potential over the course of a transition. i.e.
\begin{equation}
    F_{\Phi}(s, s') = \Phi(s') - \Phi(s).
\end{equation}
This shaping term is added to the standard reward during training: if in the original MDP $M$ the reward is $r(s, a)$ for transitioning from $s$ to $s'$ with action $a \sim \pi(s)$, then in the new MDP $M'$ the reward is updated to be
\begin{equation}
    r'(s, a) = r(s, a) +  F(s, s')
\end{equation}
at all but the last timestep, for which $r'(s_H, a_H) = r(s_H, a_H) + F(s_H, s_1)$. Observe that for any policy $\pi \in \Pi$, $J(\pi, r) = J(\pi, r')$ as the $F(s, s')$ term telescopes. Thus, $\pi^{\star} = \arg\max_{\pi \in \Pi} J(\pi, r') = \arg\max_{\pi \in \Pi} J(\pi, r)$ -- we preserve \textit{policy optimality}. Thus, in theory, PBRS only affects the speed at which we converge to the optimal policy, rather than the value of the policy we end up converging to.

\subsection{Evolution Strategies}
Evolution Strategies (ES) are zeroth-order, population-based, stochastic optimisation algorithms. First, they use random noise to generate a population of candidate solutions. These solutions are then evaluated using a fitness function. Lastly, the population is iteratively improved over time by assigning higher weight to better-performing population members. This causes the population to move closer and closer to the optimal solution, and the process is repeated until a satisfactory solution is found. Recently, ES has been successfully applied to a variety of tasks \citep{Real_Aggarwal_Huang_Le_2019, salimans2017evolution, such2018deep}. ES algorithms are gradient-free and well-suited for (meta-)optimisation problems where the objective function is noisy or non-differentiable and the search space is large or complex \citep{BEYER2000239, lange2023evosax, lu2023adversarial, lu2022discovered, houthooft2018evolved}.
This includes reward function shaping \citep{niekum} and RL hyperparameter search \citep{ELFWING20183}.

There are several types of ES algorithms, one of the most well known is the covariance matrix adaptation evolution strategy (CMA-ES) \citep{cmaes}, which
represents the population by a full-covariance multivariate Gaussian. Although CMA-ES can be applied to our problem, it has only proven successful in low to medium dimension optimisation spaces.
Another widely applied ES algorithm is OpenAI-ES \citep{salimans2017evolution} which estimates the gradient through the following function:
\[
\nabla_\theta \mathds{E}_{\epsilon \sim N(0,1)} F(\theta + \sigma \epsilon) = \frac{1}{\sigma} \mathds{E}_{\epsilon \sim N(0,1)} \{F(\theta + \sigma \epsilon)\epsilon\}
\]
This is an unbiased estimate and, in contrast to meta-gradient approaches, ES avoids the need to backpropagate the gradient through the whole training procedure, which often results in biased gradients due to truncation \citep{58337, metz2022gradients, liu2022theoretical}. As CMA-ES struggles with higher dimensional problems, we use OpenAI-ES for optimizing shaping terms across all of our experiments.

\section{Reward-Centric Challenges of Efficient IRL}

We attempt to solve the IRL Game (Eq. \ref{eq:irl-game}) via a more interaction-efficient \textit{primal} strategy: taking small steps on the reward function via gradient descent and on the policy via RL, as in \citet{ho2016generative}. As we described in the related work, in contrast to \textit{dual} algorithms that repeatedly retrain the policy at each step, this means we have weaker guarantees as far as the performance of the policy trained on the recovered reward. This is more than just a theoretical concern: as we will demonstrate in the following sections, standard GAIL-like algorithms often recover rewards that, when optimised from scratch, do not lead to strong policies. Unfortunately, switching back to a dual strategy might require an infeasible amount of interaction for the scale of problems more modern primal methods seek to solve. We therefore are left with our first open question to ponder:

\noindent \textit{\textbf{Challenge 1:} How do we ensure the final reward function returned by primal IRL methods permits effective (even if not efficient) retraining from scratch?}

Assuming that we successfully solve the \textbf{Challenge 1}, a key next question is that of \textit{interaction-efficient learning}.
Observe that because Eq. \ref{eq:irl-game} can be written as a difference of performances, it is \textit{invariant} to shaping potential-based terms. Thus, for the wide variety of algorithms that are essentially solving this game (e.g. MaxEnt IRL \citep{ziebart2008maximum} or GAIL \citep{ho2016generative} -- see \citet{swamy2021moments} for a more complete list), we have no way to ensure that they are picking well-shaped rewards, and therefore have no way to encourage the discriminator to pick rewards that are suitable for re-training. This issue is well known even in the classical IRL literature \citep{ng2000algorithms}. This leads to the next key question we seek to answer:

\noindent \textit{\textbf{Challenge 2:} How should we modify the discriminator learning process to ensure that the overall IRL procedure returns well-shaped rewards for efficient retraining?} 

In theory at least, given a reward function $r$, there is a clean example of a shaping term that permits efficient training. Consider setting $\Phi(s) = V^{\star}(s)$: the value function of the optimal policy $\pi^{\star}$ and assume deterministic dynamics. Then, by the definition of an \textit{advantage function}, we have that 
\begin{equation*}
    r'(s, a) = r(s, a) + V^{\star}(s') - V^{\star}(s) = A^{\star}(s, a).
\end{equation*}
The greedy policy with respect to this modified reward (i.e. the optimal advantage function $A^{\star}$) is $\pi^*$. Thus, we only need a planning horizon of 1 to compute the optimal policy. At heart, this is why potential-based reward shaping speeds up RL: it reduces the amount of planning the agent has to perform. In general, one has to pay exponentially in the horizon of the problem for RL \citep{kakade2003sample}, so a perfectly-shaped reward function can provide an \textit{exponential} speedup in terms of overall interaction efficiency.

In the language of search-based planning, $V^{\star}(s)$ can be thought of as an optimal \textit{heuristic}. Unfortunately, without access to $\pi^{\star}$, it is rather difficult to compute $V^{\star}$ for all but small tabular problems (e.g. via value iteration). Even with access to $\pi^{\star}$, estimating a $\hat{V}^{\star}$ that is accurate \textit{uniformly} over the state space is rather challenging, as we'd need to roll out $\pi^{\star}$ from a wide variety of states, including those entirely outside of its induced state visitation distribution. In practice, one can often at best hope to have access to the critic used in the training of $\pi^{\star}$, which is likely only accurate on states visited by $\pi^{\star}$ and not the states visited early on in training by weak policies, where shaping is most important. Thus, in practice, we are left with an open question.

\noindent \textit{\textbf{Challenge 3:} In practice, how do we learn a potential-based shaping term that is useful throughout the course of training from scratch?}

We now discuss concrete and scalable solutions to each of these challenges before validating their empirical efficacy.

\begin{algorithm*}[t]
\caption{Reward Shaping with Evolution Strategies}
\label{alg:shaping_es}
\begin{algorithmic}
\STATE \textbf{Input:} Reward function $\hat{r}$, population size $N$, inner loop updates $M$, learning rate $\alpha$, noise standard deviation $\sigma$
\STATE Initialise potential parameters $\theta$
\STATE \algcommentlight{Outer-loop, optimise shaping parameters}
\REPEAT  
\STATE Generate Gaussian noise $\epsilon_1, . . . \epsilon_N \sim \mathcal{N}(0, I)$ to generate $N$ members in the population
\FOR {$i = 1, \dots ,N$}
    \STATE $\Phi_{\theta_i} = \Phi_{\theta} + \sigma \epsilon_i$
    \STATE $AUC_i = 0$
    \STATE Initialise policy $\pi_{i 0}$
    \STATE \algcommentlight{Inner-loop, tracking RL training efficiency}
    \FOR {$j = 1, \dots ,M$}
        \STATE $\pi_{ij} \leftarrow$ RL Step on reward $\hat{r}' = F_{\Phi_{\theta_i}} + \hat{r}$
        \STATE $AUC_i \leftarrow AUC_i + \mathbb{E}_{\xi \sim \pi_{ij}}[\sum^{H-1}_{h=0} \hat{r}'(s_h, a_h, s_{h+1})]$
    \ENDFOR
\ENDFOR
\STATE \algcommentlight{Estimate gradient and update meta (shaping) parameters}
\STATE $\mathcal{L}_i \leftarrow -AUC_i$
\STATE $\theta \leftarrow \theta - \alpha \frac{1}{N\sigma} \sum^{N}_{i=1} \mathcal{L}_i \epsilon_i$
\UNTIL{convergence}
\STATE \textbf{Output:} Final shaping function $\Phi_\theta$
\end{algorithmic}
\end{algorithm*}

\begin{algorithm*}[t]
\caption{\acro{}: Evolution Strategies for Generalisable Imitation}
\label{alg:evil}
\begin{algorithmic}
\STATE \textbf{Input:} Expert trajectories $\mathcal{D}_E$,  Learning rate $\alpha$, Ensemble size $K$,
\STATE \algcommentlight{Step \#1: Run IRL++}
\STATE Initialise policy $\pi_0$, reward functions $f_0^{1:K}$, and buffer $\mathcal{D}_0$.
\FOR {$i = 1, ..., N$}
    \STATE $\mathcal{D}_{i} \leftarrow \mathcal{D}_{i-1} \cup \{\xi_i \sim \pi_i \}$
    \STATE $\ell_i(f) = \mathbb{E}_{\xi \sim \mathcal{D}_{i}}\left[\sum_h^H f(s_h, a_h) \right] - \mathbb{E}_{\xi \sim \mathcal{D}_{E}}\left[\sum_h^H f(s_h, a_h) \right] + ||f||_2$
    \STATE $\forall k \in [K]$, $f_{i+1}^k \leftarrow f_{i}^k - \alpha \nabla_f \ell_i$
    \STATE Take small RL step on $\frac{-1}{K} \sum_k^K f_i^k$ to get $\pi_{i+1}$.
\ENDFOR
\STATE \algcommentlight{Step \#2: Shape IRL reward}
\STATE Run Algorithm \ref{alg:shaping_es} on $\hat{r} = \frac{-1}{K} \sum_k^K f_N^k$ to evolve shaping term $\Phi_\theta$.
\STATE \textbf{Output:} Shaped reward $F_{\Phi_\theta} + \hat{r}$.
\end{algorithmic}
\end{algorithm*}

\section{\acro{}: Evolution Strategies for Generalisable Imitation Learning}
We now provide solutions to the preceding challenges.

\subsection{Solution 1: Improving Retrainability in IRL}
In order to improve retrainability of the rewards recovered by primal IRL algorithms, we propose a trifecta of strategies that can be applied across a variety of base algorithms.

\noindent \textbf{1A: Policy Buffer.} To mitigate the risk of the IRL discriminator forgetting valuable signals it previously provided during an earlier policy update, we maintain an ongoing buffer containing all past policy trajectories. This continuous retention allows for the consistent retraining of the IRL discriminator on the full history of learner policies. This differs from the more common practice of taking a small gradient step on the classification loss between the expert and most recent learner policy. In a sense, this can be thought of as moving from procedure reminiscent of Online Gradient Descent \citep{zinkevich2003online} to one reminiscent of Follow the Regularised Leader \citep{mcmahan2011follow}. While both are no-regret algorithms in theory, prior work in imitation learning has found that the latter can sometimes produce more stable results \citep{ross2011reduction}.

\noindent \textbf{1B: Discriminator and Policy Ensembles.} To address potential errors where the discriminator might assign an unusually high value to a state not visited by the policy, we adopt an ensemble-based approach. The policy reward is calculated as the average of all the discriminators. While ensembling techniques are common in RL for approximations of pessimism \citep{kidambi2020morel}, we instead use them to enlarge the portion of the state space where our discriminator provides useful feedback, which can be important during the random exploration that is usually a critical part of the  start of RL training. Accordingly, to ensure that each discriminator is trained on a sufficiently different set of states, we also train an ensemble of policies, using data from one policy per discriminator.

\noindent \textbf{1C: Random Policy Resets.} During IRL training, it's possible for the inner learner policy to converge prematurely, limiting exploration of all relevant states. This might cause the discriminator to overfit to a specific learner state distribution, which might differ from the totality of states seen during a fresh re-training. To elide this concern, we occasionally re-initialise the learner policy during training, with linearly decreasing probability as training advances. This still allows the learner to match expert performance while enhancing retrainability. In a sense, this can be thought of as annealing from dual to primal IRL over the course of a single training run, allowing us to reap the benefits inherent to both families of approaches.

Combined, all these modifications (along with other slight changes like improved regularisation on the discriminator) significantly improve retraining performance under the recovered reward. We call the resulting algorithm IRL++. We also performed an ablation of these components and other techniques we found to provide limited benefits -- see Figure \ref{fig:ablation_ant} in the appendix for more details.

\subsection{Solution 2: Decoupling Shaping from Discrimination}
We propose using a two-stage procedure to improve re-training efficiency for IRL: first, learning a reward, and second, learning a shaping term to be added in during the re-training procedure. This allows us to avoid the issue that sequence of loss functions that the discriminator sees during game-solving are invariant to how well shaped the chosen reward function. Critically, because shaping terms do not change the set of optimal policies \citep{ng1999policy}, optimising the recovered reward plus the shaping term cannot affect the correctness of the overall procedure and therefore preserves the strong performance guarantees of IRL.

A bit more explicitly, we propose first running IRL++ to recover a reward function that admits effective (but not efficient) retraining. Second, we optimise for a shaping term that maximises the area under the curve of the reward recovered by IRL++ (as we have no access to the ground truth reward function). This gets directly at the objective we care about -- retraining interaction efficiency -- without going through the detour of trying to learn a value function network that is accurate on the set of states encountered during retraining. The area under the curve of an RL training curve as a function of a shaping term added to the reward function is clearly a complex and non-differentiable objective. We therefore need an shaping term optimisation procedure that is amenable to such circumstances, which leads to the last component of our overall proposed procedure.

\subsection{Solution 3: ES for Potential-Based Shaping}
\begin{figure*}[t]
    \centering
    \begin{subfigure}[b][][t]{0.24\textwidth}
    \label{subfig:2a}
    \includegraphics[width=\textwidth]{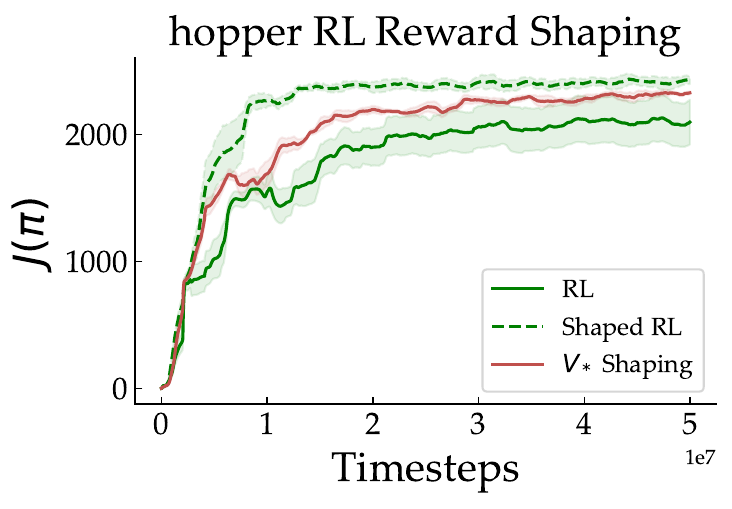}
    \end{subfigure}
    \begin{subfigure}[b][][t]{0.24\textwidth}
    \label{subfig:2b}
    \includegraphics[width=\textwidth]{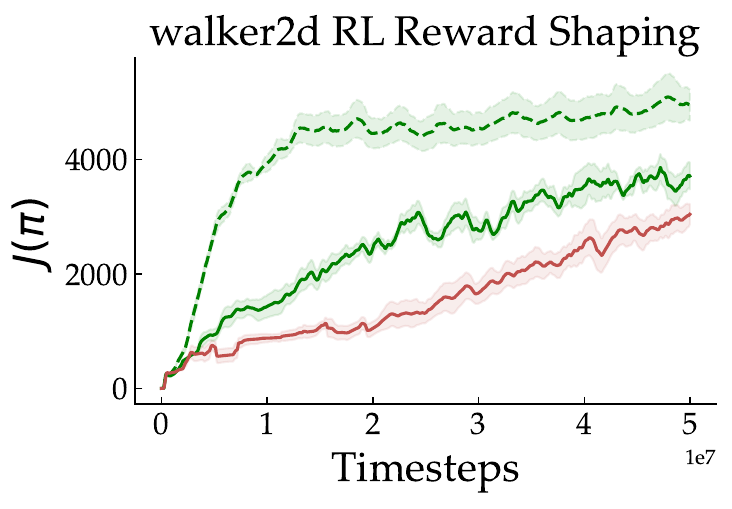}
    \end{subfigure}
    \begin{subfigure}[b][][t]{0.24\textwidth}
    \label{subfig:2c}
    \includegraphics[width=\textwidth]{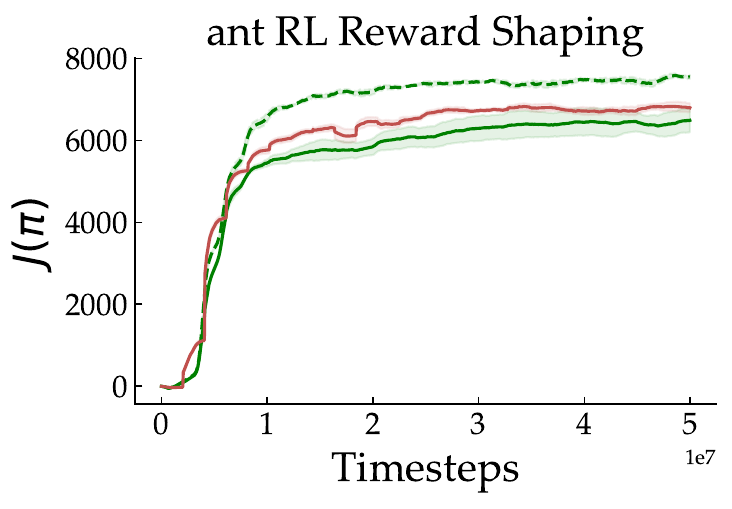}
    \end{subfigure}
    \begin{subfigure}[b][][t]{0.24\textwidth}
    \label{subfig:2d}
    \includegraphics[width=\textwidth]{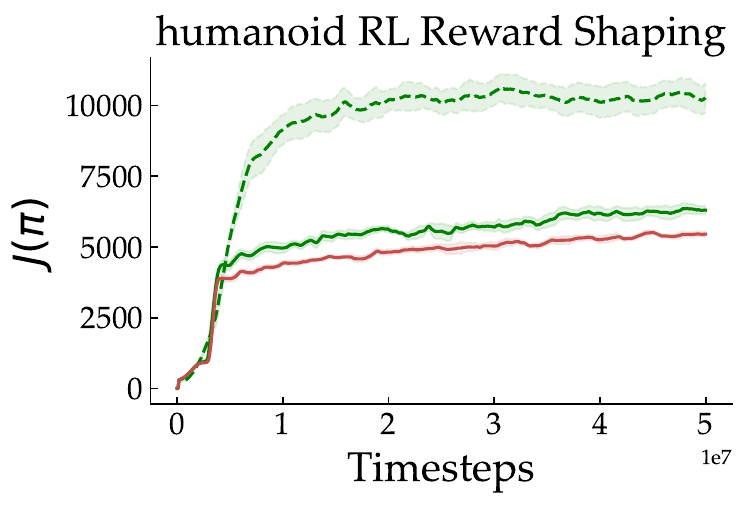}
    \end{subfigure}
    \caption{\textbf{Shaping RL.} Our method successfully recovers a potential-based reward function that helps an RL agent learn faster. We compare against the baseline of using the expert value function as a shaping term when retraining and non-shaped RL training on the real reward. We compute standard error across 5 seeds. $J(\pi)$ is the performance of the learner under the ground truth reward.}
    \label{fig:rl_and_shaped}
\end{figure*}

Rather than attempting to learn $V^{\star}$, we propose simply \textit{evolving} a potential $\Phi$ that leads to faster training. Specifically, we use the \textit{area under the curve} (AUC) of the performance $J(\pi, \hat{r})$ vs. environment interactions plot as the \textit{fitness function} for the ES optimisation. As shown in Algorithm \ref{alg:shaping_es}, we calculate reward AUC during training by saving the policy's performance after each gradient update. After the inner loop procedure is complete, the AUC is passed onto the OpenES algorithm. This estimates the gradient to maximise AUC and updates the shaping meta parameters accordingly. We note that such a technique is of interest in learning shaping terms even for vanilla RL with known rewards and confirm its efficacy at doing so in Figure \ref{fig:rl_and_shaped}.

We refer to the combination of IRL++ with an evolved shaping term as \acro{}: \textbf{EVolution strategies for Imitation Learning}. See Algs. \ref{alg:shaping_es} and \ref{alg:evil} for full details of our method.

\subsubsection{When is \acro{} a good idea?}
A natural question when reading the preceding section might be the following: \textit{if we care about improving the sample efficiency of IRL retraining, doesn't expending a large interaction budget by repeatedly retraining the agent from scratch to learn a good shaping term via evolution seem counter-intuitive?} Indeed, if one were to perform such shaping-via-evolution in the real world, \acro{} would likely cost more to implement than it would save in the final retraining in the novel (or same) environment. 

However, for a variety of domains, we have access to a reasonably accurate simulator. Such simulators are often used to perform the interactive learning component of IRL, with the recovered reward then optimised in the real world to recover strong policies / trajectories (e.g. in the autonomous driving domain \citep{vinitsky2022nocturne, gulino2024waymax}). As discussed above, this recovered reward can be (and often is in practice) poorly shaped. Thus, \textit{by running \acro{} inside the simulator, we can trade exploration in the real world for exploration in simulation, which is likely to be cheaper and safer.} Of course, then one has to contend with the well-known sim2real gap, both in terms of the recovered reward and on the learned shaping term. We investigate this concern in depth in our experiments section, where we explore the effect of a shaping term in improving retraining performance in a novel environment (e.g. the real world instead of the simulator the shaping term was trained in) and find encouraging results across the board. Thus, we believe that on problems where a cheap approximate simulator exists, \acro{} presents a feasible path for improving the interaction efficiency of retraining on the reward recovered by IRL. Even on problems where this is not the case, one could potentially \textit{learn} a model to perform IRL in, as explored by \citet{ren2023hyrbid}. It would be an interesting direction for future work to perform ES-based shaping inside a learned model (where interaction is relatively inexpensive) and see how much benefit it provides for real-world retraining.

\begin{figure*}[t]
    \centering
    \begin{subfigure}[b][][t]{0.24\textwidth}
    \label{subfig:3a}
    \includegraphics[width=\textwidth]{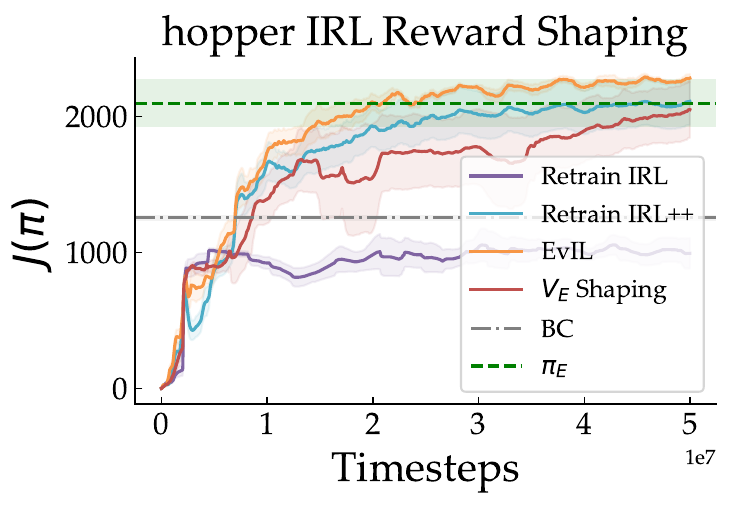}
    \end{subfigure}
    \begin{subfigure}[b][][t]{0.24\textwidth}
    \label{subfig:3b}
    \includegraphics[width=\textwidth]{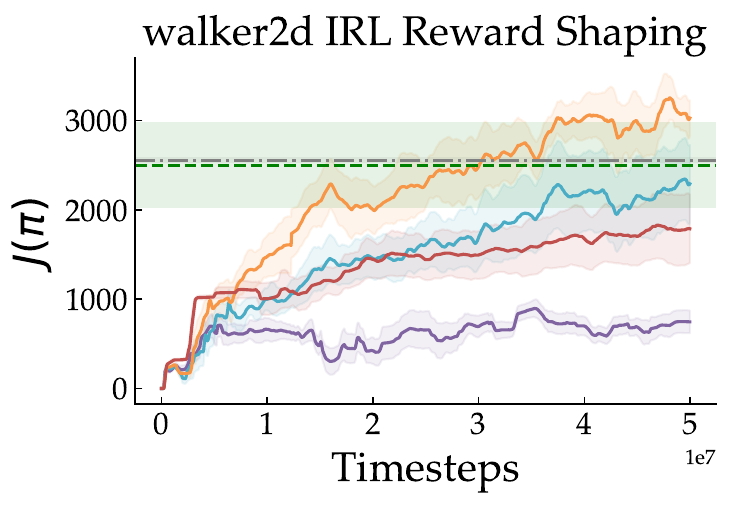}
    \end{subfigure}
    \begin{subfigure}[b][][t]{0.24\textwidth}
    \label{subfig:3c}
    \includegraphics[width=\textwidth]{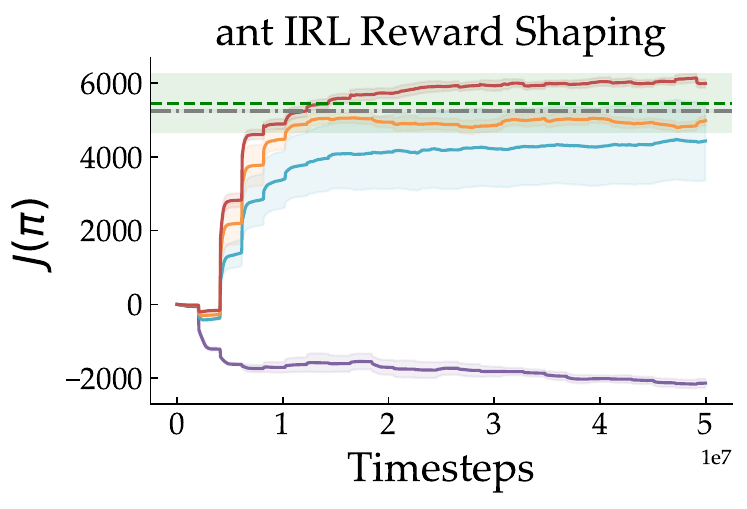}
    \end{subfigure}
    \begin{subfigure}[b][][t]{0.24\textwidth}
    \label{subfig:3d}
    \includegraphics[width=\textwidth]{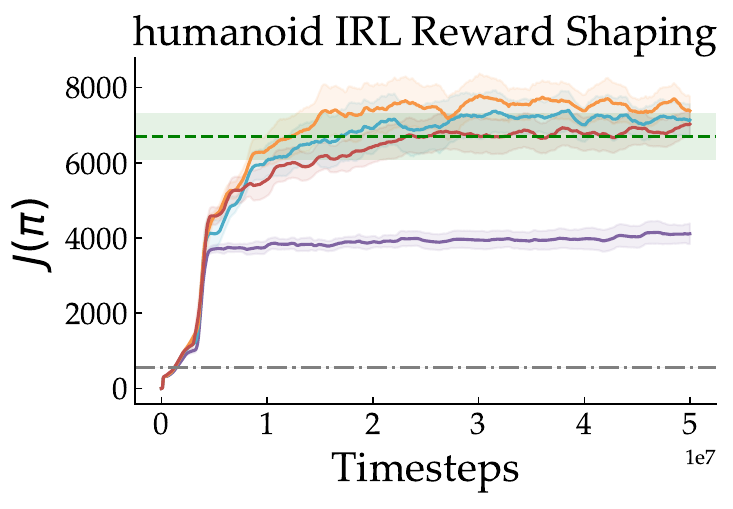}
    \end{subfigure}
    \caption{\textbf{Shaping IRL.} We show our method can successfully recover a potential-based reward function that makes the recovered reward function easier to learn. We use the shaping term combined with a reward recovered from IRL++. We compare against three baselines: the reward recovered from an ensemble of discriminators, without shaping, the reward recovered by a classic IRL method, and the IRL++ reward shaped using the expert value function when retraining. For each, we train on 5 seeds, with shading representing standard error.}
    \label{fig:irl_and_shaped}
\end{figure*}

\begin{figure*}[t]
    \centering
    \begin{subfigure}[b][][t]{0.24\textwidth}
    \label{subfig:4a}
    \includegraphics[width=\textwidth]{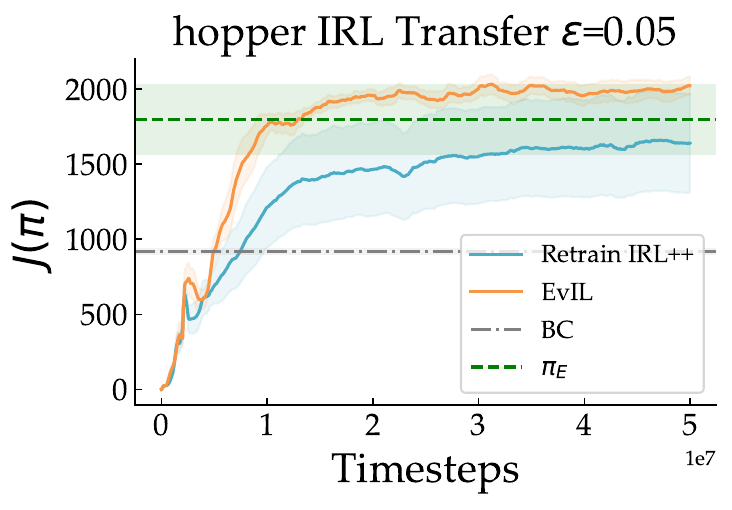}
    \end{subfigure}
    \begin{subfigure}[b][][t]{0.24\textwidth}
    \label{subfig:4b}
    \includegraphics[width=\textwidth]{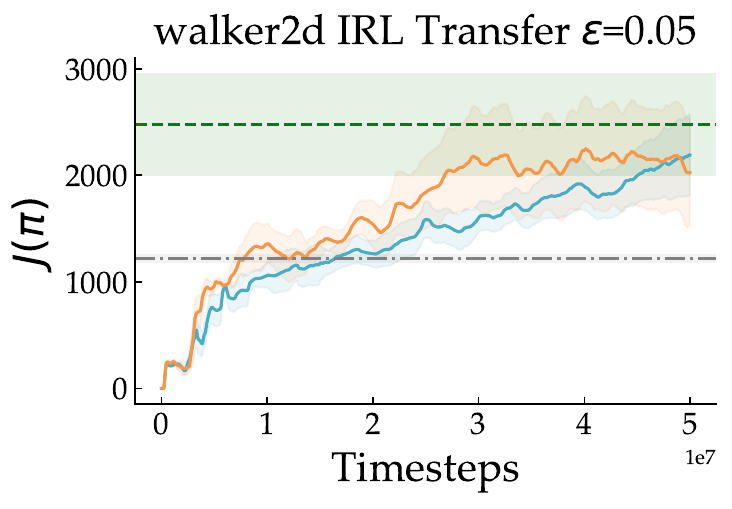}
    \end{subfigure}
    \begin{subfigure}[b][][t]{0.24\textwidth}
    \label{subfig:4c}
    \includegraphics[width=\textwidth]{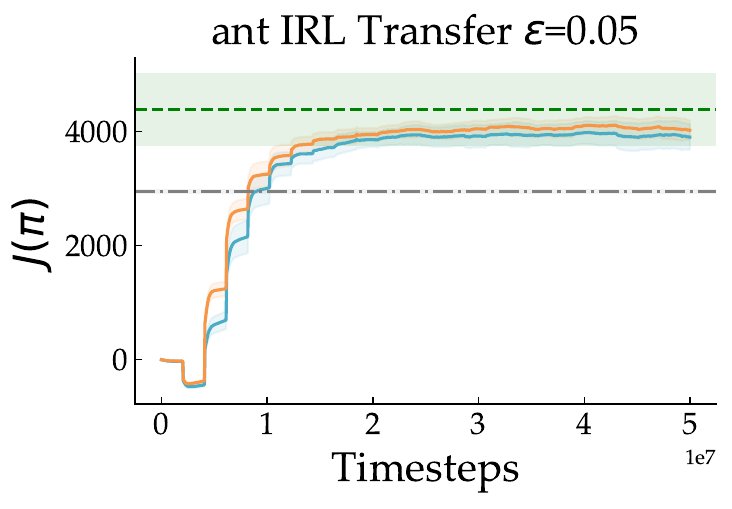}
    \end{subfigure}
    \begin{subfigure}[b][][t]{0.24\textwidth}
    \label{subfig:4d}
    \includegraphics[width=\textwidth]{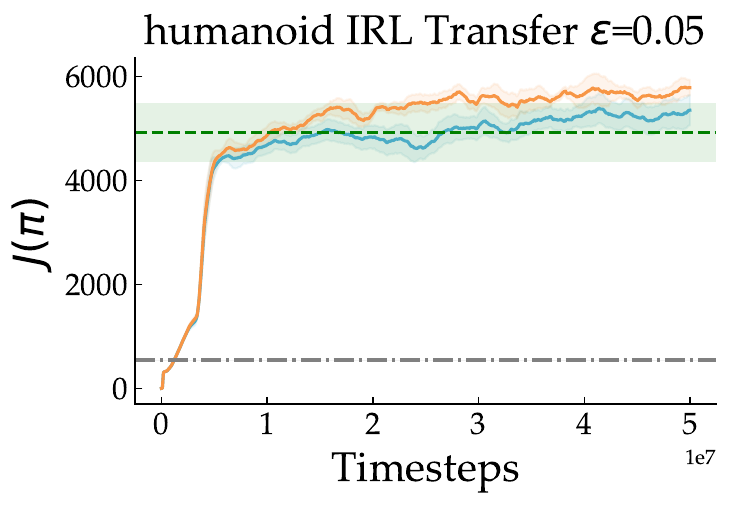}
    \end{subfigure}
    \caption{\textbf{\acro{} Transfer on Trembling Hand Environment} \acro{} outperforms both BC and IRL++ on transfer to an environment where, with $\epsilon$ probability, a random action is executed in the environment rather than the one the agent selected. IRL++ out-performs BC, highlighting the importance of interactive training for effective transfer.}
    \label{fig:tremble_irl_and_shaped}
\end{figure*}

\begin{figure*}[t]
    \centering
    \begin{subfigure}[b][][t]{0.24\textwidth}
    \label{subfig:5a}
    \includegraphics[width=\textwidth]{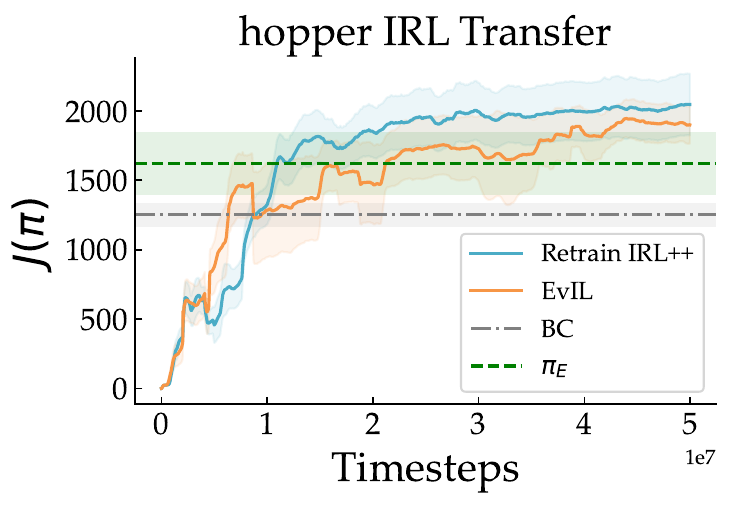}
    \end{subfigure}
    \begin{subfigure}[b][][t]{0.24\textwidth}
    \label{subfig:5b}
    \includegraphics[width=\textwidth]{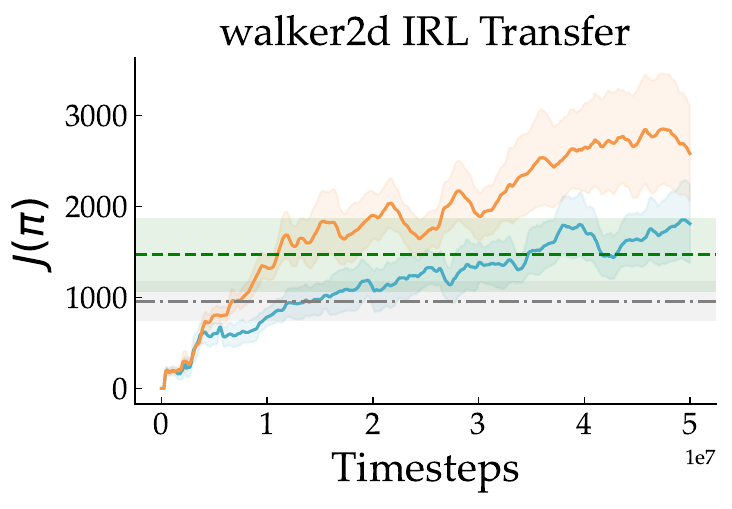}
    \end{subfigure}
    \begin{subfigure}[b][][t]{0.24\textwidth}
    \label{subfig:5c}
    \includegraphics[width=\textwidth]{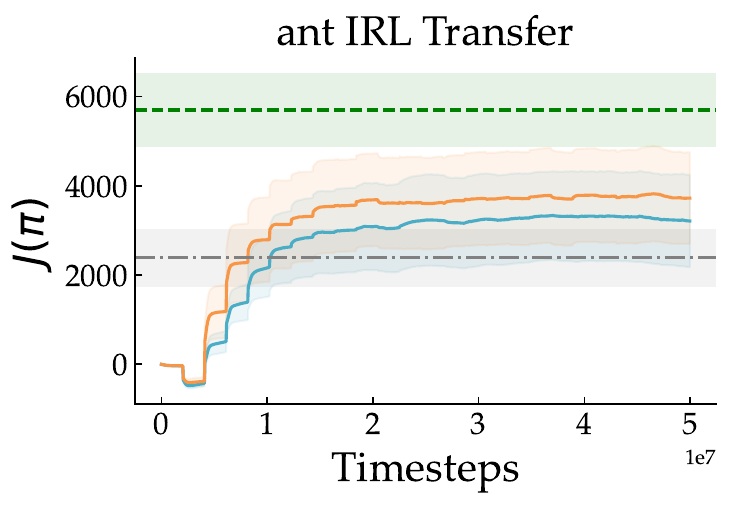}
    \end{subfigure}
    \begin{subfigure}[b][][t]{0.24\textwidth}
    \label{subfig:5d}
    \includegraphics[width=\textwidth]{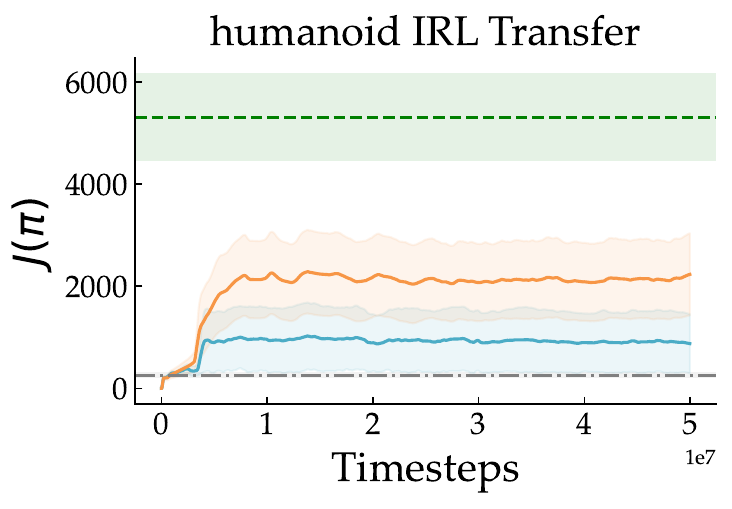}
    \end{subfigure}
    \caption{\textbf{\acro{} Transfer on Randomised Dynamics Environment} \acro{} outperforms both BC and IRL++ on transfer to an environment where link lengths and joint ranges are randomly sampled and differ from the demonstrations. As before, IRL++ out-performs BC.}
    \label{fig:transfer_irl_and_shaped}
\end{figure*}

\section{Experimental Results} \label{section:experiments}
In this section, we aim to answer the following questions:
\begin{enumerate}
    \item \textbf{Can ES (Algorithm \ref{alg:shaping_es}) be used to learn a shaping function that improves interaction efficiency?} We first investigate whether ES successfully learns an effective shaping function in RL with a hand-designed reward. We compare to a strong baseline: the critic of a policy trained extensively on the ground-truth reward.
    \item \textbf{Does our modified IRL procedure ensure we recover reward functions that admit effective retraining?} We investigate whether, by taking the final reward function returned by our modified procedure and optimising it extensively in the training environment, we are able to find a policy with similar performance to the expert.
    \item \textbf{By combining the preceding techniques (i.e. \acro{}), are we able to efficiently and effectively retrain policies in both the source and target environments?} We investigate whether we are able to more sample-efficiently find high quality policies, both in the source and in target environments. Specifically, we consider environments where we add stochasticity to the dynamics or slightly change the link lengths of our agent.
\end{enumerate}

Together these questions get at our over-arching goal with \acro{}: \textit{the ability to robustly transfer expert behaviour to novel environments with limited interaction.}

We conduct our experiments across three distinct MuJoCo environments: Hopper, Walker, and Ant. All learners receive 100 trajectories from the expert policy, trained using Proximal Policy Optimisation (PPO) \citep{schulman2017proximal} over 5e7 timesteps. We perform two sets of transfer experiments per environment. In the first, with probability $p_{tremble}$, a randomly selected action is executed instead of the one determined by the agent's policy, potentially leading the agent to encounter previously unseen states at test time. In the second, more challenging set of experiments, we randomly sample link lengths from a distribution with significant support outside of the dynamics the expert demonstrations were collected under.

Our IRL implementation is built upon the GAIL framework proposed in \citet{ho2016generative}. To enhance its performance, we follow the recommendations of \citet{swamy2022minimax}, such as incorporating a gradient penalty \citep{gulrajani2017improved} to stabilise the discriminator training process and a linear learning rate decay for the actor, critic, and discriminator. Within the inner loop of our optimisation strategy, we employ PPO \citep{schulman2017proximal}. Across all experiments, our IRL discriminator is trained on states only, rather than state-action pairs. This reflects an under-appreciated benefit of IRL: the ability to perform imitation without access to action labels, which are difficult to obtain in various domains (e.g. learning from third-person videos).

All our code is implemented in JAX \citep{jax2018github} using the PureJaxRL \cite{lu2022discovered}, Brax \citep{brax2021github}, and evosax \cite{lange2023evosax} libraries to maximise parallelisation of training. For all our experiments, we use 5 parallel learners and 5 discriminators. We release the code we used for all of our experiments at \textbf{\texttt{\url{https://github.com/SilviaSapora/evil}}}.

\subsection{ES-Based Shaping of Ground-Truth Rewards}
As a sanity check, we first confirm that performing ES-based shaping recovers $V^{\star}$ in a small tabular in Figure \ref{fig:heatmap_gridworld} in the appendix. Next, in Figure \ref{fig:rl_and_shaped}, we compare the performance of an RL agent trained with our evolved shaping function to the performance of an agent trained on the ground truth reward. We see that we are consistently able to recover a shaping that leads to better interaction efficiency, even on top of a hand-designed reward function. Furthermore, we find that we consistently compete with or exceed the performance of using the critic of a strong policy as a learned shaping term. Recall that this is a very strong baseline that assumes access to a privileged piece of information. We attribute the strength of our method to directly optimising for the desired objective, allowing one to ensure one does not waste representational capacity on accurately fitting $V^{\star}$ on states that are unimportant for efficient re-training.

In greater detail, we evolve the shaping term over 300 generations with a population size of 64. To expedite the optimisation procedure, we only shaped the first 50\% of RL training for all environments. We note that none of our training runs had converged after 300 generations, indicating we could have kept training to further improve the efficacy of our learned shaping term, indicating the results we report are a lower bound on the maximal efficacy of our method.

\subsection{IRL++: Retraining with the Recovered Reward}
In Figure \ref{fig:irl_and_shaped}, we show that naively re-training on the final IRL reward function performs poorly. Specifically, in the Ant environment, employing this reward function results in a negative reward, while in the Hopper, Walker and Humanoid environments, the policy plateaus, ceasing to learn at a sub-optimal performance. These policies might be stuck in local minimum and never converge, regardless of how many environment interactions they get to experience. In contrast, after employing the suggestions from the preceding section (i.e. IRL++), we find that retraining from scratch (in teal) gives us significantly better performance with enough training time. This confirms that IRL++ permits effective re-training. However, in environments like Walker and Ant, training is perhaps less efficient than desirable, which segues into the results for our complete method.

\subsection{\acro{}: Retraining with an Evolved Shaping Term}
For \acro{}, we combine our proposed reward model ensembling and policy resets (i.e. IRL++) with an evolved shaping term. The shaping rewards were evolved over 300 generations with a population size of 64. The Hopper and Walker shaping function were optimised on the first 20\% of training only, while Ant and Humanoid doubled the amount of training considered for the last 100 generations. In Figure \ref{fig:irl_and_shaped}, we see that across the board, \acro{} (in orange) leads to marked improvements over optimising the unshaped reward. For both Hopper and Humanoid, we are able to match / exceed expert performance while for Ant we are able to achieve a score thousands higher than the IRL baseline. The $V_E$ shaping baseline we consider is particularly strong due to the privileged information it has access to: an estimate expert's value function (i.e. the final critic from the RL training we used to generate the expert). Interestingly, we find that the performance of $V_E$ shaping is inconsistent: it hinders training in the Hopper, Walker and Humanoid environments, but is effective in the Ant environment. We hypothesise this is due to differing levels of covariate shift in terms of the state distribution of the initial and expert policies. %

\subsection{\acro{} Transfer Performance}
To assess the transferability of the policies learned by all methods, we consider two sets of modifications to the environment. First, we consider, with probability $p_{tremble} = 0.05$, forcing the policy to execute a random action, a corruption that is not present in the expert demonstrations. We retrain under the recovered reward in this stochastic environment but do \textbf{\textit{not}} give the learner access to the test environment during the inverse RL procedure. Echoing our argument in the introduction of the paper, we find that the interactive methods are able to more effectively transfer than offline approaches like behavioural cloning. Furthermore, we find that our learned shaping term improves both the \textit{effectiveness} (final performance) and \textit{efficiency} (interaction before performance peaks) of the retraining procedure.

To highlight the robustness of our method, we then consider a significantly more challenging transfer task: controlling an agent with different link lengths than the one the expert demonstrations were collected with. Specifically, we introduce random variations in the link lengths and joint ranges of the target agents (details postponed to the Appendix). Performance was evaluated by sampling 10 different agent variations from each MuJoCo environment.
Our findings in Figure \ref{fig:transfer_irl_and_shaped} show that \acro{} consistently outperforms the BC and unshaped baselines. BC performance exhibits a notable decline across all three environments when compared to both the trembling hand case and the standard environment, highlighting the difficulty of this transfer learning task. 

In summary, our experiments support our hypothesis that the combination of ingredients that make up \acro{} are critical for ensuring that we are able to \textit{effectively} and \textit{efficiently} mimic expert behaviour in novel environments.

\section{Conclusion}
\textbf{Summary. } We empirically demonstrate that the reward functions returned by modern deep IRL algorithms do not admit effective retraining. We propose a set of fixes, IRL++, that help recover reward functions that, when optimised, lead to policies that match expert performance. However, such retraining can be rather expensive in terms of the number of interactions required. In response, we propose \acro{}: a full-stack algorithm that combines IRL++ with an evolution strategies-based approach for learning shaping terms. On top of the retraining efficacy IRL++ gives us, \acro{} gives us efficiency. We validate \acro{}'s performance on challenging transfer tasks involving unseen environments and find that it is able to out-perform the prior art.

\section*{Acknowledgements}
SS was supported by Google TPU Research Cloud (TRC) and Google Cloud Research Credits program. 

GKS is supported by his family and friends and thanks Drew Bagnell for his thoughts on our core algorithm and suggestion of the value function shaping baseline. 

JF is partially funded by the UKI grant EP/Y028481/1 (originally selected for funding by the ERC).
JF is also supported by the JPMC Research Award and the Amazon Research Award.

\section*{Impact Statement}
This paper presents work whose goal is to advance the field of Machine Learning. There are many potential societal consequences of our work, none which we feel must be specifically highlighted here.

\section*{Contribution Statements}
\begin{itemize}
    \item \textbf{SS} implemented all the code, performed all experiments, released a high-quality open-source implementation of the proposed methods and proposed most of the key components of IRL++.
    \item \textbf{GS} came up with the core algorithmic idea of using evolution to learn a potential-based shaping term, assisted with debugging the inverse RL-based component of the experiments, and wrote most of the paper.
    \item \textbf{CL} initially proposed the project and assisted with debugging the evolution-based component of the experiments.
    \item \textbf{YWT} and \textbf{JNF} advised the project. \textbf{JNF} helped write parts of the paper.
\end{itemize}

\bibliography{main}

\begin{thebibliography}{63}
\providecommand{\natexlab}[1]{#1}
\providecommand{\url}[1]{\texttt{#1}}
\expandafter\ifx\csname urlstyle\endcsname\relax
  \providecommand{\doi}[1]{doi: #1}\else
  \providecommand{\doi}{doi: \begingroup \urlstyle{rm}\Url}\fi

\bibitem[Abbeel \& Ng(2004)Abbeel and Ng]{abbeel2004apprenticeship}
Abbeel, P. and Ng, A.~Y.
\newblock Apprenticeship learning via inverse reinforcement learning.
\newblock In \emph{Proceedings of the twenty-first international conference on Machine learning}, pp.\ ~1, 2004.

\bibitem[Barnes et~al.(2023)Barnes, Abueg, Lange, Deeds, Trader, Molitor, Wulfmeier, and O'Banion]{barnes2023massively}
Barnes, M., Abueg, M., Lange, O.~F., Deeds, M., Trader, J., Molitor, D., Wulfmeier, M., and O'Banion, S.
\newblock Massively scalable inverse reinforcement learning in google maps.
\newblock \emph{arXiv preprint arXiv:2305.11290}, 2023.

\bibitem[Beyer(2000)]{BEYER2000239}
Beyer, H.-G.
\newblock Evolutionary algorithms in noisy environments: theoretical issues and guidelines for practice.
\newblock \emph{Computer Methods in Applied Mechanics and Engineering}, 186\penalty0 (2):\penalty0 239--267, 2000.
\newblock ISSN 0045-7825.
\newblock \doi{https://doi.org/10.1016/S0045-7825(99)00386-2}.
\newblock URL \url{https://www.sciencedirect.com/science/article/pii/S0045782599003862}.

\bibitem[Bradbury et~al.(2018)Bradbury, Frostig, Hawkins, Johnson, Leary, Maclaurin, Necula, Paszke, Vander{P}las, Wanderman-{M}ilne, and Zhang]{jax2018github}
Bradbury, J., Frostig, R., Hawkins, P., Johnson, M.~J., Leary, C., Maclaurin, D., Necula, G., Paszke, A., Vander{P}las, J., Wanderman-{M}ilne, S., and Zhang, Q.
\newblock {JAX}: composable transformations of {P}ython+{N}um{P}y programs, 2018.
\newblock URL \url{http://github.com/google/jax}.

\bibitem[Cooke et~al.(2023)Cooke, Klyne, Zhang, Laidlaw, Tambe, and Doshi-Velez]{cooke2023toward}
Cooke, L.~H., Klyne, H., Zhang, E., Laidlaw, C., Tambe, M., and Doshi-Velez, F.
\newblock Toward computationally efficient inverse reinforcement learning via reward shaping.
\newblock \emph{arXiv preprint arXiv:2312.09983}, 2023.

\bibitem[Elfwing et~al.(2018)Elfwing, Uchibe, and Doya]{ELFWING20183}
Elfwing, S., Uchibe, E., and Doya, K.
\newblock Sigmoid-weighted linear units for neural network function approximation in reinforcement learning.
\newblock \emph{Neural Networks}, 107:\penalty0 3--11, 2018.
\newblock ISSN 0893-6080.
\newblock \doi{https://doi.org/10.1016/j.neunet.2017.12.012}.
\newblock URL \url{https://www.sciencedirect.com/science/article/pii/S0893608017302976}.
\newblock Special issue on deep reinforcement learning.

\bibitem[Finn et~al.(2016)Finn, Christiano, Abbeel, and Levine]{finn2016connection}
Finn, C., Christiano, P., Abbeel, P., and Levine, S.
\newblock A connection between generative adversarial networks, inverse reinforcement learning, and energy-based models, 2016.

\bibitem[Flajolet et~al.(2022)Flajolet, Monroc, Beguir, and Pierrot]{flajolet2022fast}
Flajolet, A., Monroc, C.~B., Beguir, K., and Pierrot, T.
\newblock Fast population-based reinforcement learning on a single machine.
\newblock In \emph{International Conference on Machine Learning}, pp.\  6533--6547. PMLR, 2022.

\bibitem[Freeman et~al.(2021)Freeman, Frey, Raichuk, Girgin, Mordatch, and Bachem]{brax2021github}
Freeman, C.~D., Frey, E., Raichuk, A., Girgin, S., Mordatch, I., and Bachem, O.
\newblock Brax - a differentiable physics engine for large scale rigid body simulation, 2021.
\newblock URL \url{http://github.com/google/brax}.

\bibitem[Fu et~al.(2018)Fu, Luo, and Levine]{fu2018learning}
Fu, J., Luo, K., and Levine, S.
\newblock Learning robust rewards with adversarial inverse reinforcement learning, 2018.

\bibitem[Goodfellow et~al.(2014)Goodfellow, Pouget-Abadie, Mirza, Xu, Warde-Farley, Ozair, Courville, and Bengio]{goodfellow2014generative}
Goodfellow, I., Pouget-Abadie, J., Mirza, M., Xu, B., Warde-Farley, D., Ozair, S., Courville, A., and Bengio, Y.
\newblock Generative adversarial nets.
\newblock \emph{Advances in neural information processing systems}, 27, 2014.

\bibitem[Gulino et~al.(2024)Gulino, Fu, Luo, Tucker, Bronstein, Lu, Harb, Pan, Wang, Chen, et~al.]{gulino2024waymax}
Gulino, C., Fu, J., Luo, W., Tucker, G., Bronstein, E., Lu, Y., Harb, J., Pan, X., Wang, Y., Chen, X., et~al.
\newblock Waymax: An accelerated, data-driven simulator for large-scale autonomous driving research.
\newblock \emph{Advances in Neural Information Processing Systems}, 36, 2024.

\bibitem[Gulrajani et~al.(2017)Gulrajani, Ahmed, Arjovsky, Dumoulin, and Courville]{gulrajani2017improved}
Gulrajani, I., Ahmed, F., Arjovsky, M., Dumoulin, V., and Courville, A.
\newblock Improved training of wasserstein gans, 2017.

\bibitem[Hansen \& Ostermeier(2001)Hansen and Ostermeier]{cmaes}
Hansen, N. and Ostermeier, A.
\newblock Completely derandomized self-adaptation in evolution strategies.
\newblock \emph{Evolutionary Computation}, 9\penalty0 (2):\penalty0 159--195, 2001.
\newblock \doi{10.1162/106365601750190398}.

\bibitem[Ho \& Ermon(2016)Ho and Ermon]{ho2016generative}
Ho, J. and Ermon, S.
\newblock Generative adversarial imitation learning, 2016.

\bibitem[Houthooft et~al.(2018)Houthooft, Chen, Isola, Stadie, Wolski, Ho, and Abbeel]{houthooft2018evolved}
Houthooft, R., Chen, R.~Y., Isola, P., Stadie, B.~C., Wolski, F., Ho, J., and Abbeel, P.
\newblock Evolved policy gradients, 2018.

\bibitem[Jackson et~al.(2023)Jackson, Lu, Kirsch, Lange, Whiteson, and Foerster]{jackson2023discovering}
Jackson, M.~T., Lu, C., Kirsch, L., Lange, R.~T., Whiteson, S., and Foerster, J.~N.
\newblock Discovering temporally-aware reinforcement learning algorithms.
\newblock In \emph{Second Agent Learning in Open-Endedness Workshop}, 2023.

\bibitem[Jaderberg et~al.(2019)Jaderberg, Czarnecki, Dunning, Marris, Lever, Castañeda, Beattie, Rabinowitz, Morcos, Ruderman, Sonnerat, Green, Deason, Leibo, Silver, Hassabis, Kavukcuoglu, and Graepel]{Jaderberg_2019}
Jaderberg, M., Czarnecki, W.~M., Dunning, I., Marris, L., Lever, G., Castañeda, A.~G., Beattie, C., Rabinowitz, N.~C., Morcos, A.~S., Ruderman, A., Sonnerat, N., Green, T., Deason, L., Leibo, J.~Z., Silver, D., Hassabis, D., Kavukcuoglu, K., and Graepel, T.
\newblock Human-level performance in 3d multiplayer games with population-based reinforcement learning.
\newblock \emph{Science}, 364\penalty0 (6443):\penalty0 859–865, May 2019.
\newblock ISSN 1095-9203.
\newblock \doi{10.1126/science.aau6249}.
\newblock URL \url{http://dx.doi.org/10.1126/science.aau6249}.

\bibitem[Kakade(2003)]{kakade2003sample}
Kakade, S.~M.
\newblock \emph{On the sample complexity of reinforcement learning}.
\newblock University of London, University College London (United Kingdom), 2003.

\bibitem[Khan et~al.(2023)Khan, Willi, Kwan, Tacchetti, Lu, Grefenstette, Rockt{\"a}schel, and Foerster]{khan2023scaling}
Khan, A., Willi, T., Kwan, N., Tacchetti, A., Lu, C., Grefenstette, E., Rockt{\"a}schel, T., and Foerster, J.
\newblock Scaling opponent shaping to high dimensional games.
\newblock \emph{arXiv preprint arXiv:2312.12568}, 2023.

\bibitem[Kidambi et~al.(2020)Kidambi, Rajeswaran, Netrapalli, and Joachims]{kidambi2020morel}
Kidambi, R., Rajeswaran, A., Netrapalli, P., and Joachims, T.
\newblock Morel: Model-based offline reinforcement learning.
\newblock \emph{Advances in neural information processing systems}, 33:\penalty0 21810--21823, 2020.

\bibitem[Kitani et~al.(2012)Kitani, Ziebart, Bagnell, and Hebert]{kitani2012activity}
Kitani, K.~M., Ziebart, B.~D., Bagnell, J.~A., and Hebert, M.
\newblock Activity forecasting.
\newblock In \emph{Computer Vision--ECCV 2012: 12th European Conference on Computer Vision, Florence, Italy, October 7-13, 2012, Proceedings, Part IV 12}, pp.\  201--214. Springer, 2012.

\bibitem[Kolter et~al.(2008)Kolter, Rodgers, and Ng]{kolter2008control}
Kolter, J.~Z., Rodgers, M.~P., and Ng, A.~Y.
\newblock A control architecture for quadruped locomotion over rough terrain.
\newblock In \emph{2008 IEEE International Conference on Robotics and Automation}, pp.\  811--818. IEEE, 2008.

\bibitem[Laidlaw et~al.(2023)Laidlaw, Russell, and Dragan]{laidlaw2023bridging}
Laidlaw, C., Russell, S., and Dragan, A.
\newblock Bridging rl theory and practice with the effective horizon.
\newblock \emph{arXiv preprint arXiv:2304.09853}, 2023.

\bibitem[Lange(2023)]{lange2023evosax}
Lange, R.~T.
\newblock evosax: Jax-based evolution strategies.
\newblock In \emph{Proceedings of the Companion Conference on Genetic and Evolutionary Computation}, pp.\  659--662, 2023.

\bibitem[Likhachev et~al.(2003)Likhachev, Gordon, and Thrun]{likhachev2003ara}
Likhachev, M., Gordon, G.~J., and Thrun, S.
\newblock Ara*: Anytime a* with provable bounds on sub-optimality.
\newblock \emph{Advances in neural information processing systems}, 16, 2003.

\bibitem[Likhachev et~al.(2005)Likhachev, Stentz, and Thrun]{likhachev2005anytime}
Likhachev, M., Stentz, A., and Thrun, S.
\newblock Anytime dynamic a*: An anytime, replanning algorithm.
\newblock 2005.

\bibitem[Liu et~al.(2022)Liu, Feng, Ren, Mai, Zhu, Zhang, Wang, and Yang]{liu2022theoretical}
Liu, B., Feng, X., Ren, J., Mai, L., Zhu, R., Zhang, H., Wang, J., and Yang, Y.
\newblock A theoretical understanding of gradient bias in meta-reinforcement learning, 2022.

\bibitem[Lu et~al.(2022{\natexlab{a}})Lu, Kuba, Letcher, Metz, Schroeder~de Witt, and Foerster]{lu2022discovered}
Lu, C., Kuba, J., Letcher, A., Metz, L., Schroeder~de Witt, C., and Foerster, J.
\newblock Discovered policy optimisation.
\newblock \emph{Advances in Neural Information Processing Systems}, 35:\penalty0 16455--16468, 2022{\natexlab{a}}.

\bibitem[Lu et~al.(2022{\natexlab{b}})Lu, Willi, De~Witt, and Foerster]{lu2022model}
Lu, C., Willi, T., De~Witt, C. A.~S., and Foerster, J.
\newblock Model-free opponent shaping.
\newblock In \emph{International Conference on Machine Learning}, pp.\  14398--14411. PMLR, 2022{\natexlab{b}}.

\bibitem[Lu et~al.(2023)Lu, Willi, Letcher, and Foerster]{lu2023adversarial}
Lu, C., Willi, T., Letcher, A., and Foerster, J.~N.
\newblock Adversarial cheap talk.
\newblock In \emph{International Conference on Machine Learning}, pp.\  22917--22941. PMLR, 2023.

\bibitem[Lupu et~al.(2024)Lupu, Lu, Liesen, Lange, and Foerster]{lupu2024behaviour}
Lupu, A., Lu, C., Liesen, J.~L., Lange, R.~T., and Foerster, J.~N.
\newblock Behaviour distillation.
\newblock In \emph{The Twelfth International Conference on Learning Representations}, 2024.
\newblock URL \url{https://openreview.net/forum?id=qup9xD8mW4}.

\bibitem[McMahan(2011)]{mcmahan2011follow}
McMahan, B.
\newblock Follow-the-regularized-leader and mirror descent: Equivalence theorems and l1 regularization.
\newblock In \emph{Proceedings of the Fourteenth International Conference on Artificial Intelligence and Statistics}, pp.\  525--533. JMLR Workshop and Conference Proceedings, 2011.

\bibitem[Metz et~al.(2022)Metz, Freeman, Schoenholz, and Kachman]{metz2022gradients}
Metz, L., Freeman, C.~D., Schoenholz, S.~S., and Kachman, T.
\newblock Gradients are not all you need, 2022.

\bibitem[Ng et~al.(1999)Ng, Harada, and Russell]{ng1999policy}
Ng, A.~Y., Harada, D., and Russell, S.
\newblock Policy invariance under reward transformations: Theory and application to reward shaping.
\newblock In \emph{Icml}, volume~99, pp.\  278--287. Citeseer, 1999.

\bibitem[Ng et~al.(2000)Ng, Russell, et~al.]{ng2000algorithms}
Ng, A.~Y., Russell, S., et~al.
\newblock Algorithms for inverse reinforcement learning.
\newblock In \emph{Icml}, volume~1, pp.\ ~2, 2000.

\bibitem[Ng et~al.(2006)Ng, Coates, Diel, Ganapathi, Schulte, Tse, Berger, and Liang]{ng2006autonomous}
Ng, A.~Y., Coates, A., Diel, M., Ganapathi, V., Schulte, J., Tse, B., Berger, E., and Liang, E.
\newblock Autonomous inverted helicopter flight via reinforcement learning.
\newblock In \emph{Experimental robotics IX}, pp.\  363--372. Springer, 2006.

\bibitem[Niekum et~al.(2010)Niekum, Barto, and Spector]{niekum}
Niekum, S., Barto, A., and Spector, L.
\newblock Genetic programming for reward function search.
\newblock \emph{Autonomous Mental Development, IEEE Transactions on}, 2:\penalty0 83 -- 90, 07 2010.
\newblock \doi{10.1109/TAMD.2010.2051436}.

\bibitem[Pomerleau(1988)]{pomerleau1988alvinn}
Pomerleau, D.~A.
\newblock Alvinn: An autonomous land vehicle in a neural network.
\newblock \emph{Advances in neural information processing systems}, 1, 1988.

\bibitem[Puterman(2014)]{puterman2014markov}
Puterman, M.~L.
\newblock \emph{Markov decision processes: discrete stochastic dynamic programming}.
\newblock John Wiley \& Sons, 2014.

\bibitem[Ratliff et~al.(2009)Ratliff, Silver, and Bagnell]{ratliff2009learning}
Ratliff, N.~D., Silver, D., and Bagnell, J.~A.
\newblock Learning to search: Functional gradient techniques for imitation learning.
\newblock \emph{Autonomous Robots}, 27\penalty0 (1):\penalty0 25--53, 2009.

\bibitem[Real et~al.(2019)Real, Aggarwal, Huang, and Le]{Real_Aggarwal_Huang_Le_2019}
Real, E., Aggarwal, A., Huang, Y., and Le, Q.~V.
\newblock Regularized evolution for image classifier architecture search.
\newblock \emph{Proceedings of the AAAI Conference on Artificial Intelligence}, 33\penalty0 (01):\penalty0 4780--4789, Jul. 2019.
\newblock \doi{10.1609/aaai.v33i01.33014780}.
\newblock URL \url{https://ojs.aaai.org/index.php/AAAI/article/view/4405}.

\bibitem[Ren et~al.(2023)Ren, Swamy, Wu, Bagnell, and Choudhury]{ren2023hyrbid}
Ren, J., Swamy, G., Wu, Z.~S., Bagnell, J.~A., and Choudhury, S.
\newblock Hybrid inverse reinforcement learning.
\newblock 2023.
\newblock URL \url{https://www.robot-learning.ml/2023/files/paper42.pdf}.

\bibitem[Ross et~al.(2011)Ross, Gordon, and Bagnell]{ross2011reduction}
Ross, S., Gordon, G., and Bagnell, D.
\newblock A reduction of imitation learning and structured prediction to no-regret online learning.
\newblock In \emph{Proceedings of the fourteenth international conference on artificial intelligence and statistics}, pp.\  627--635. JMLR Workshop and Conference Proceedings, 2011.

\bibitem[Russell \& Norvig(2010)Russell and Norvig]{russell2010artificial}
Russell, S.~J. and Norvig, P.
\newblock \emph{Artificial intelligence a modern approach}.
\newblock 2010.

\bibitem[Rutherford et~al.(2023)Rutherford, Ellis, Gallici, Cook, Lupu, Ingvarsson, Willi, Khan, de~Witt, Souly, et~al.]{rutherford2023jaxmarl}
Rutherford, A., Ellis, B., Gallici, M., Cook, J., Lupu, A., Ingvarsson, G., Willi, T., Khan, A., de~Witt, C.~S., Souly, A., et~al.
\newblock Jaxmarl: Multi-agent rl environments in jax.
\newblock \emph{arXiv preprint arXiv:2311.10090}, 2023.

\bibitem[Salimans et~al.(2017)Salimans, Ho, Chen, Sidor, and Sutskever]{salimans2017evolution}
Salimans, T., Ho, J., Chen, X., Sidor, S., and Sutskever, I.
\newblock Evolution strategies as a scalable alternative to reinforcement learning, 2017.

\bibitem[Schulman et~al.(2017)Schulman, Wolski, Dhariwal, Radford, and Klimov]{schulman2017proximal}
Schulman, J., Wolski, F., Dhariwal, P., Radford, A., and Klimov, O.
\newblock Proximal policy optimization algorithms, 2017.

\bibitem[Silver et~al.(2010)Silver, Bagnell, and Stentz]{silver2010learning}
Silver, D., Bagnell, J.~A., and Stentz, A.
\newblock Learning from demonstration for autonomous navigation in complex unstructured terrain.
\newblock \emph{The International Journal of Robotics Research}, 29\penalty0 (12):\penalty0 1565--1592, 2010.

\bibitem[Such et~al.(2018)Such, Madhavan, Conti, Lehman, Stanley, and Clune]{such2018deep}
Such, F.~P., Madhavan, V., Conti, E., Lehman, J., Stanley, K.~O., and Clune, J.
\newblock Deep neuroevolution: Genetic algorithms are a competitive alternative for training deep neural networks for reinforcement learning, 2018.

\bibitem[Swamy et~al.(2021)Swamy, Choudhury, Bagnell, and Wu]{swamy2021moments}
Swamy, G., Choudhury, S., Bagnell, J.~A., and Wu, S.
\newblock Of moments and matching: A game-theoretic framework for closing the imitation gap.
\newblock In \emph{International Conference on Machine Learning}, pp.\  10022--10032. PMLR, 2021.

\bibitem[Swamy et~al.(2022)Swamy, Rajaraman, Peng, Choudhury, Bagnell, Wu, Jiao, and Ramchandran]{swamy2022minimax}
Swamy, G., Rajaraman, N., Peng, M., Choudhury, S., Bagnell, J., Wu, S.~Z., Jiao, J., and Ramchandran, K.
\newblock Minimax optimal online imitation learning via replay estimation.
\newblock \emph{Advances in Neural Information Processing Systems}, 35:\penalty0 7077--7088, 2022.

\bibitem[Swamy et~al.(2023)Swamy, Choudhury, Bagnell, and Wu]{swamy2023inverse}
Swamy, G., Choudhury, S., Bagnell, J.~A., and Wu, Z.~S.
\newblock Inverse reinforcement learning without reinforcement learning, 2023.

\bibitem[Syed \& Schapire(2007)Syed and Schapire]{syed2007game}
Syed, U. and Schapire, R.~E.
\newblock A game-theoretic approach to apprenticeship learning.
\newblock \emph{Advances in neural information processing systems}, 20, 2007.

\bibitem[Tiapkin et~al.(2023)Tiapkin, Belomestny, Calandriello, Moulines, Naumov, Perrault, Valko, and Menard]{tiapkin2023regularized}
Tiapkin, D., Belomestny, D., Calandriello, D., Moulines, E., Naumov, A., Perrault, P., Valko, M., and Menard, P.
\newblock Regularized rl.
\newblock \emph{arXiv preprint arXiv:2310.17303}, 2023.

\bibitem[Vinitsky et~al.(2022)Vinitsky, Lichtl{\'e}, Yang, Amos, and Foerster]{vinitsky2022nocturne}
Vinitsky, E., Lichtl{\'e}, N., Yang, X., Amos, B., and Foerster, J.
\newblock Nocturne: a scalable driving benchmark for bringing multi-agent learning one step closer to the real world.
\newblock \emph{Advances in Neural Information Processing Systems}, 35:\penalty0 3962--3974, 2022.

\bibitem[Werbos(1990)]{58337}
Werbos, P.
\newblock Backpropagation through time: what it does and how to do it.
\newblock \emph{Proceedings of the IEEE}, 78\penalty0 (10):\penalty0 1550--1560, 1990.
\newblock \doi{10.1109/5.58337}.

\bibitem[Wu \& Tian(2017)Wu and Tian]{Wu2017Doom}
Wu, Y. and Tian, Y.
\newblock Training agent for first-person shooter game with actor-critic curriculum learning.
\newblock 2017.

\bibitem[Ziebart et~al.(2012)Ziebart, Dey, and Bagnell]{ziebart2012probabilistic}
Ziebart, B., Dey, A., and Bagnell, J.~A.
\newblock Probabilistic pointing target prediction via inverse optimal control.
\newblock In \emph{Proceedings of the 2012 ACM international conference on Intelligent User Interfaces}, pp.\  1--10, 2012.

\bibitem[Ziebart et~al.(2008{\natexlab{a}})Ziebart, Maas, Bagnell, Dey, et~al.]{ziebart2008maximum}
Ziebart, B.~D., Maas, A.~L., Bagnell, J.~A., Dey, A.~K., et~al.
\newblock Maximum entropy inverse reinforcement learning.
\newblock In \emph{Aaai}, volume~8, pp.\  1433--1438. Chicago, IL, USA, 2008{\natexlab{a}}.

\bibitem[Ziebart et~al.(2008{\natexlab{b}})Ziebart, Maas, Dey, and Bagnell]{ziebart2008navigate}
Ziebart, B.~D., Maas, A.~L., Dey, A.~K., and Bagnell, J.~A.
\newblock Navigate like a cabbie: Probabilistic reasoning from observed context-aware behavior.
\newblock In \emph{Proceedings of the 10th international conference on Ubiquitous computing}, pp.\  322--331, 2008{\natexlab{b}}.

\bibitem[Zinkevich(2003)]{zinkevich2003online}
Zinkevich, M.
\newblock Online convex programming and generalized infinitesimal gradient ascent.
\newblock In \emph{Proceedings of the 20th international conference on machine learning (icml-03)}, pp.\  928--936, 2003.

\bibitem[Zucker et~al.(2011)Zucker, Ratliff, Stolle, Chestnutt, Bagnell, Atkeson, and Kuffner]{zucker2011optimization}
Zucker, M., Ratliff, N., Stolle, M., Chestnutt, J., Bagnell, J.~A., Atkeson, C.~G., and Kuffner, J.
\newblock Optimization and learning for rough terrain legged locomotion.
\newblock \emph{The International Journal of Robotics Research}, 30\penalty0 (2):\penalty0 175--191, 2011.

\end{thebibliography}
\bibliographystyle{icml2024}

\newpage
\appendix
\onecolumn
\section{Appendix}

\begin{figure*}[htbp]
    \centering
    \begin{subfigure}[t]{0.24\textwidth}
    \label{subfig:irL_a}
    \includegraphics[width=\textwidth]{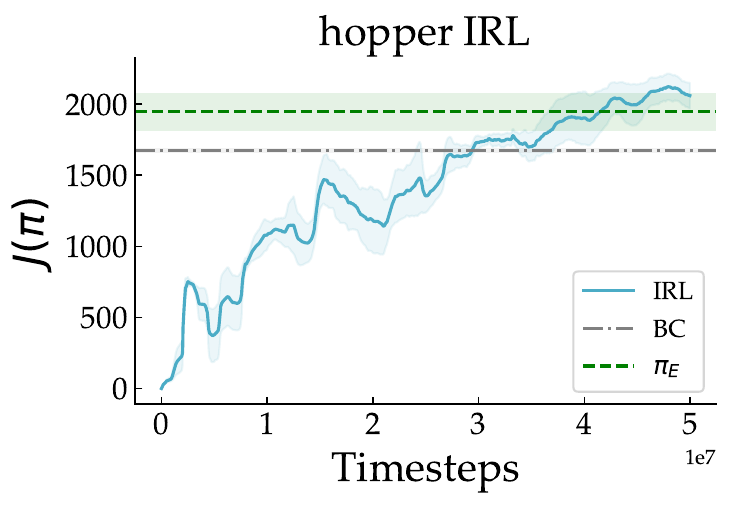}
    \end{subfigure}
    \begin{subfigure}[t]{0.24\textwidth}
    \label{subfig:irl_b}
    \includegraphics[width=\textwidth]{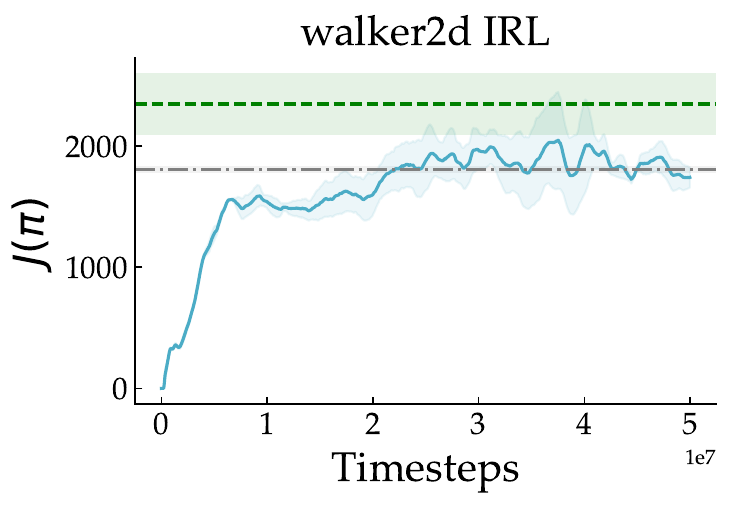}
    \end{subfigure}
    \begin{subfigure}[t]{0.24\textwidth}
    \label{subfig:irl_c}
    \includegraphics[width=\textwidth]{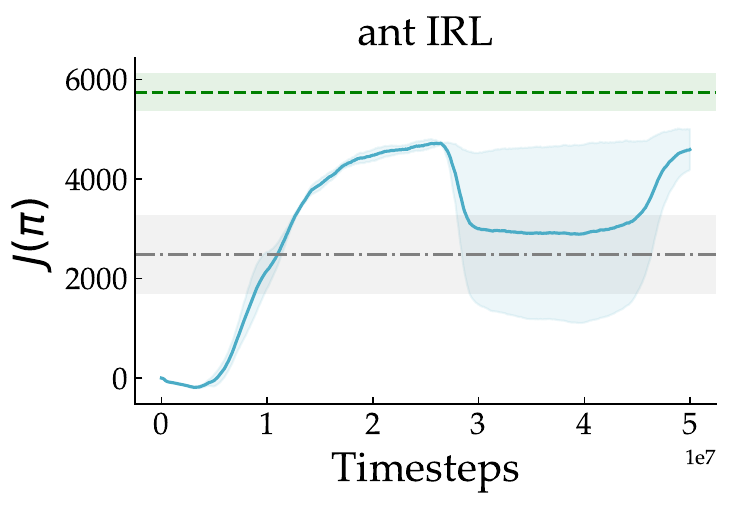}
    \end{subfigure}
    \begin{subfigure}[t]{0.24\textwidth}
    \label{subfig:irl_d}
    \includegraphics[width=\textwidth]{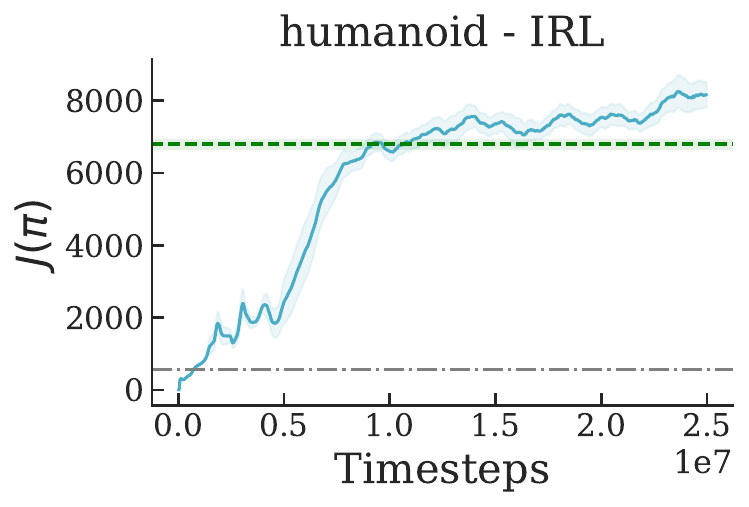}
    \end{subfigure}
    \caption{\textbf{IRL} IRL training across MuJoCo environments compared to expert and BC performance. 5 seeds each, shading represents standard error.}
    \label{fig:irl_train}
\end{figure*}

\begin{figure*}[htbp]
    \centering
    \begin{subfigure}[t]{0.26\textwidth}
    \label{subfig:ab_a}
    \includegraphics[width=\textwidth]{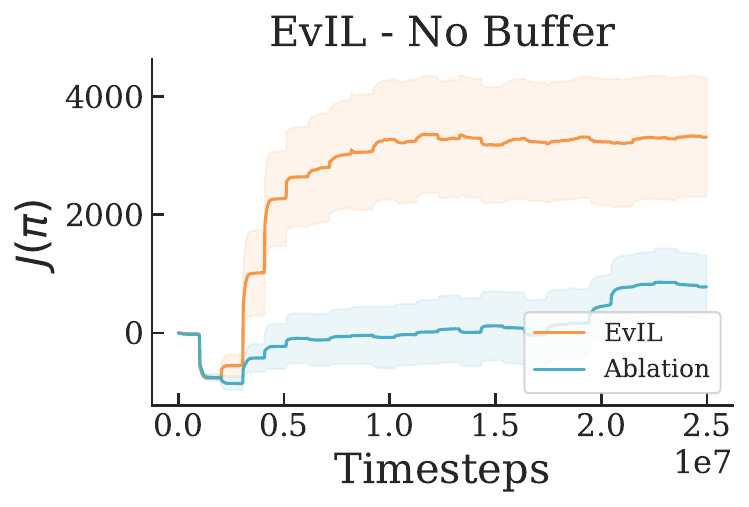}
    \end{subfigure}
    \begin{subfigure}[t]{0.26\textwidth}
    \label{subfig:ab_b}
    \includegraphics[width=\textwidth]{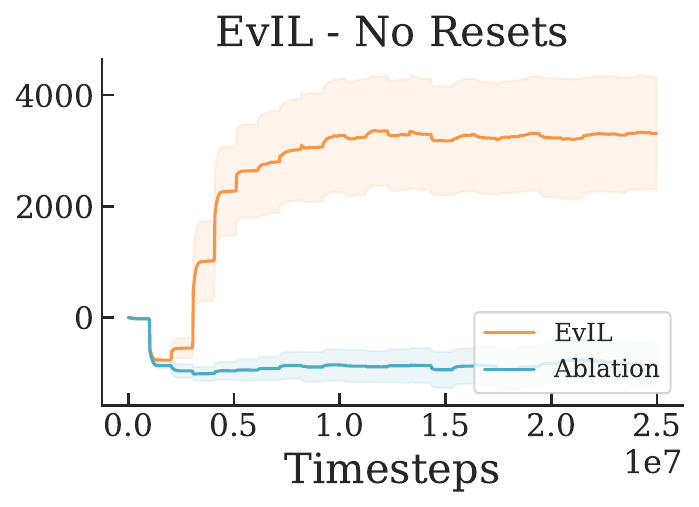}
    \end{subfigure}
    \begin{subfigure}[t]{0.26\textwidth}
    \label{subfig:ab_c}
    \includegraphics[width=\textwidth]{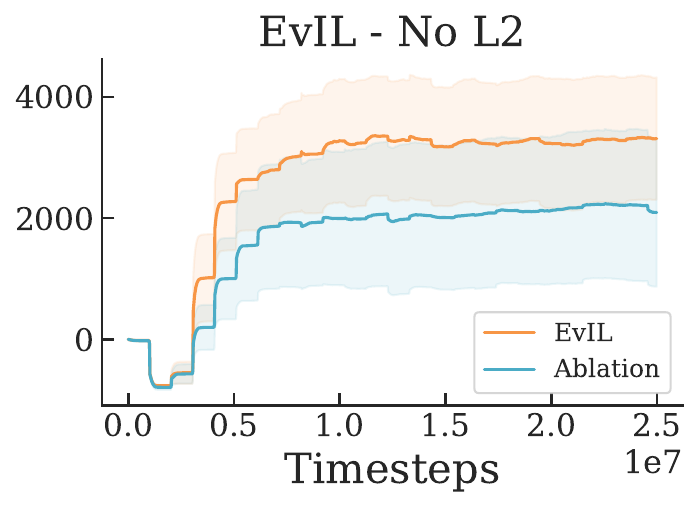}
    \end{subfigure}
    \begin{subfigure}[t]{0.26\textwidth}
    \label{subfig:ab_d}
    \includegraphics[width=\textwidth]{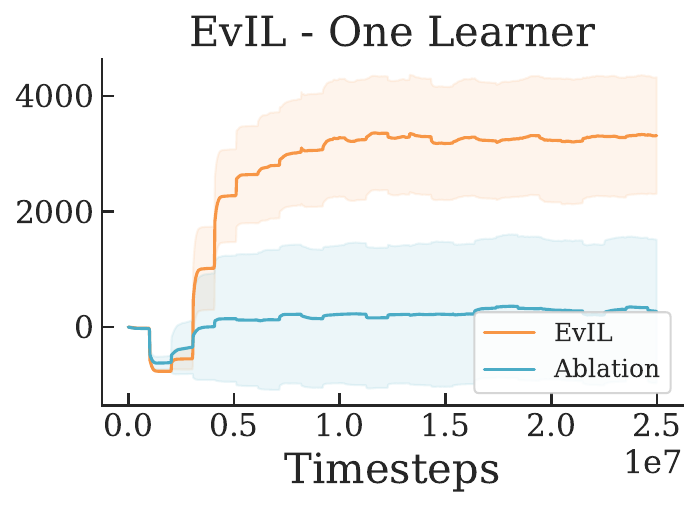}
    \end{subfigure}
    \begin{subfigure}[t]{0.26\textwidth}
    \label{subfig:ab_e}
    \includegraphics[width=\textwidth]{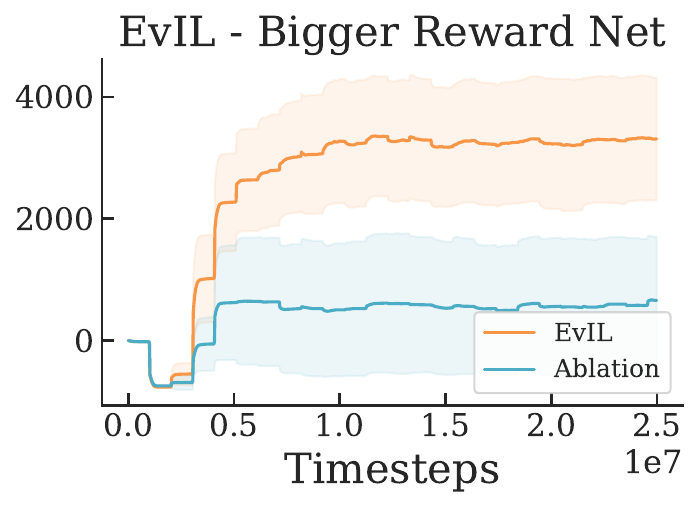}
    \end{subfigure}
    \caption{Ablation for \acro{} on the Ant environment. \acro{} uses an ensemble of size 5. \textbf{a}. We compare \acro{} with an ensemble that does not utilise a replay buffer. In this case, the discriminator only observes samples from the most recent policy version. \textbf{b}.  We compare against a version of \acro{} that doesn't implement resets of the learner policies. \textbf{c}. We check the effect of $\ell_2$ regularisation on the reward network \textbf{d}. The ablation explores the effect of using a single learner seed (and 5 discriminators) instead of multiple seeds. \textbf{e}. We use a bigger reward network (two hidden layers of size 512) to verify the issue doesn't lie in our reward network not having enough representational capacity to accurately describe the expert behaviour.}
    \label{fig:ablation_ant}
\end{figure*}

\begin{figure*}[htbp]
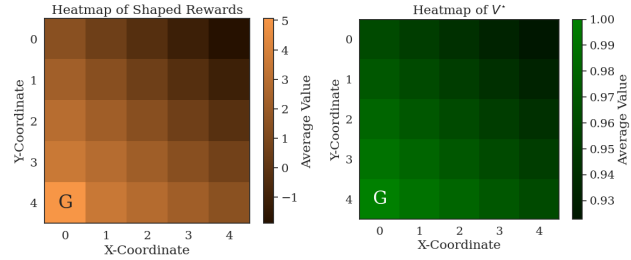

    \centering
    \begin{subfigure}[t]{0.24\textwidth}
    \label{subfig:gridworld_a}
    \includegraphics[width=\textwidth]{./plots/heatmap_gridworld_shaped.pdf}
    \end{subfigure}
    \begin{subfigure}[t]{0.24\textwidth}
    \label{subfig:gridworld_b}
    \includegraphics[width=\textwidth]{./plots/gridworld_heatmap_value.pdf}
    \end{subfigure}
    \caption{Heatmaps representing the values of the shaped reward (left) and the optimal value function (right) corresponding to each position the agent could move to in the 5x5 grid. The goal (represented with the G) is in the bottom left corner of the grid.}
    \label{fig:heatmap_gridworld}
\end{figure*}

\begin{table}[]
\centering
\begin{tabular}{|p{4cm}|p{3cm}|p{3cm}|}
\cline{1-3}
\multicolumn{1}{|c|}{\textbf{Method}} & \multicolumn{2}{|c|}{\textbf{Number of Agents}} \\ \cline{2-3}
 & \textbf{1 Agent} & \textbf{5 Agents} \\ \hline
\texttt{EvIL}                     & 0.4082 ($\pm$ 0.1166)  & \textbf{0.6696} ($\pm$ 0.0176)  \\ \hline
\texttt{EvIL} - No buffer         & 0.2702 ($\pm$ 0.1204) & 0.5739 ($\pm$ 0.0192) \\ \hline
\texttt{EvIL} - No L2             & 0.3942 ($\pm$ 0.0861) & 0.6124 ($\pm$ 0.0233)  \\ \hline
\texttt{EvIL} - Bigger Reward Net & 0.2603 ($\pm$ 0.0901) & 0.5058 ($\pm$ 0.0057)  \\ \hline
\texttt{EvIL} - No Reset          & 0.0951 ($\pm$ 0.1111) & 0.2544 ($\pm$ 0.0170)  \\ \hline
IRL                      & -0.0180 ($\pm$ 0.0895) & 0.1424 ($\pm$ 0.0153)  \\ \hline
IRL + Resets             & 0.2912 ($\pm$ 0.1573) & 0.6277  ($\pm$ 0.0279) \\ \hline
IRL + L2                 & -0.0579 ($\pm$ 0.2612) & 0.2724 ($\pm$ 0.1020)  \\ \hline
IRL + Bigger Reward Net  & -0.0940 ($\pm$ 0.1415) & -0.0216 ($\pm$ 0.0444) \\ \hline
IRL + Buffer             & 0.2053 ($\pm$ 0.0413) & 0.4310 ($\pm$ 0.0180) \\ \hline
\end{tabular}
\caption{Correlation between the recovered reward and the real reward at the end of the IRL procedure. Ablation for \acro{} and IRL. We report mean and standard error on the states visited by the learner.}
\end{table}

\section{Environment Dynamics Randomisation}
To vary the source tasks, we sample the links’ size and the joint range parameters as follows:
\begin{enumerate}
  \item Hopper:
        \begin{itemize}
            \item Foot Link Size: $U[0.05, 0.07] \times U[0.1755, 0.2145]$
            \item Leg Link Size: $U[0.03, 0.05] \times U[0.23, 0.27]$
            \item Thigh Link Size: $U[0.04, 0.06] \times U[0.2, 0.25]$
            \item Torso Link Size: $U[0.04, 0.06] \times U[0.2, 0.25]$
            \item Thigh Joint $U[-2.79253, -2.44346] \times U[0, 0]$
        \end{itemize}
  \item Walker:
        \begin{itemize}
            \item Left Foot Link Size: $U[0.05, 0.07] \times U[0.09, 0.11]$
            \item Right Foot Link Size: $U[0.05, 0.07] \times U[0.09, 0.11]$
            \item Left Leg Link Size: $U[0.03, 0.05] \times U[0.23, 0.27]$
            \item Right Leg Link Size: $U[0.03, 0.05] \times U[0.23, 0.27]$
            \item Left Thigh Link Size: $U[0.04, 0.06] \times U[0.2, 0.25]$
            \item Right Thigh Link Size: $U[0.04, 0.06] \times U[0.2, 0.25]$
            \item Left Thigh Joint $U[-2.79253, -2.44346] \times U[0, 0]$
        \end{itemize}
  \item Ant:
        \begin{itemize}
            \item Left Leg Link Size: $U[0.05, 0.11]$
            \item Right Leg Link Size: $U[0.05, 0.11]$
            \item Left Leg Link Size: $U[0.05, 0.11]$
            \item Back Right Leg Link Size: $U[0.05, 0.11]$
            \item Aux 1 Link Size: $U[0.05, 0.11]$
            \item Aux 2 Link Size: $U[0.05, 0.11]$
            \item Aux 3 Link Size: $U[0.05, 0.11]$
            \item Aux 4 Link Size: $U[0.05, 0.11]$
            \item Left Ankle Link Size: $U[0.05, 0.11]$
            \item Right Ankle Link Size: $U[0.05, 0.11]$
            \item Back Left Ankle Link Size: $U[0.05, 0.11]$
            \item Back Right Ankle Link Size: $U[0.05, 0.11]$
            \item Joint Hip 1 $U[-40, -20] \times U[20, 40]$
            \item Joint Hip 2 $U[-40, -20] \times U[20, 40]$
            \item Joint Hip 3 $U[-40, -20] \times U[20, 40]$
            \item Joint Hip 4 $U[-40, -20] \times U[20, 40]$
            \item Joint Ankle 1 $U[-80, -60] \times U[-40, -20]$
            \item Joint Ankle 2 $U[-80, -60] \times U[-40, -20]$
            \item Joint Ankle 3 $U[-80, -60] \times U[-40, -20]$
            \item Joint Ankle 4 $U[-80, -60] \times U[-40, -20]$
            
        \end{itemize}
\end{enumerate}

\section{Hyperparameters}

\begin{table}[hbt!]
\centering
\caption{Hyperparameters for Training IRL}
\label{tab:lpo_hyperparams}
\begin{tabular}{l|l}
Parameter & Value \\ \hline
Number of Reward Hidden Layers & 2 \\
Size of Reward Hidden Layer & 128 \\
Reward Activation & tanh \\
Number of Outer Loop Steps & 2441 \\
Inner Loop Learning Rate & 4e-3 \\
Inner Loop Gradient Updates & 1 \\
Inner Loop Num Steps & 10 \\
Inner Loop Num Envs & 2048 \\
Num Trajectories Sampled each Policy Update & 10 \\
Gradient Penalty Coefficient & 10 \\
Outer Loop Learning Rate Schedule & Linear \\
Number of Discriminators & 5 \\
$\ell_2$ coefficient & 0.0 \\
Outer Loop Learning Rate Start & 1e-2 \\
Outer Loop Learning Rate End & 1e-5 \\
\end{tabular}
\end{table}
For the Ant environment, the Final Outer Loop Learning Rate is 1e-6.
\begin{table}[hbt!]
\centering
\caption{Important parameters for Training Reward Shaping with ES}
\label{tab:evil_hyperparams}
\begin{tabular}{l|l}
Parameter & Value \\ \hline
Population Size & 64 \\
Number of Reward Hidden Layers & 2 \\
Size of Reward Hidden Layer & 128 \\
Reward Activation & tanh \\
Number of Generations & 600 \\
ES Sigma Init &  0.03 \\
ES Sigma Decay & 1.00 \\
ES LR & 1e-3 \\
Outer Loop Learning Rate & 5e-3 \\
\end{tabular}
\end{table}

\end{document}